\documentclass[review]{elsarticle}

\usepackage{amsmath}
\usepackage{subcaption}
\usepackage{booktabs}
\usepackage{tabularx}
\usepackage{dirtree}
\usepackage{tikz}
\usepackage{threeparttable}
\usepackage{hyperref}
\usepackage[edges]{forest}
\let\today\relax
\makeatletter
\def\ps@pprintTitle{%
    \let\@oddhead\@empty
    \let\@evenhead\@empty
    \def\@oddfoot{\footnotesize\itshape
         {Accepted manuscript for Automation in Construction. Submitted: 16 Jul 2024. Accepted: 17 Jan 2025}\hfill\today}%
    \let\@evenfoot\@oddfoot
    }
\makeatother

\begin{document}
\begin{frontmatter}
    \title{
    Segmentation Dataset for Reinforced Concrete Construction
    }
    \author[1]{Patrick Schmidt\corref{cor1}}
    \ead{pasch@dtu.dk}
    
    \author[1]{Lazaros Nalpantidis}
    \ead{lanalpa@dtu.dk}
    
    \affiliation[1]{organization={DTU - Technical University of Denmark, Department of Electrical and Photonics Engineering}, 
    addressline={Elektrovej Bygning 326}, 
    postcode={2800},
    city={Kongens Lyngby}, 
    country={Denmark}}
    
    \cortext[cor1]{Corresponding author}   

\begin{abstract}
    This paper provides a dataset of 14,805 RGB images with segmentation labels for autonomous robotic inspection of reinforced concrete defects. Baselines for the YOLOv8L-seg, DeepLabV3, and U-Net segmentation models are established. Labelling inconsistencies are addressed statistically, and their influence on model performance is analyzed. An error identification tool is employed to examine the error modes of the models. The paper demonstrates that YOLOv8L-seg performs best, achieving a validation mIOU score of up to 0.59. Label inconsistencies were found to have a negligible effect on model performance, while the inclusion of more data improved the performance. False negatives were identified as the primary failure mode. The results highlight the importance of data availability for the performance of deep learning-based models. The lack of publicly available data is identified as a significant contributor to false negatives. To address this, the paper advocates for an increased open-source approach within the construction community.
\end{abstract}

\begin{keyword}
    dataset \sep
    construction robotics \sep
    segmentation \sep
    rebar detection \sep
    shotcrete \sep
    digitization
\end{keyword}
\end{frontmatter}

%%%%%%%%%%%%%%%%%%%%%%%%%%%%%%%%%%%%%%%%%%
\section{Introduction}
The construction industry will be facing a big challenge in the upcoming years. In the European Union (EU), the number of people employed in the construction industry has steadily increased since 2015 \cite{Eurostat_2022}. However, so did the number of hours worked per person, except for a little, probably COVID-related dip in 2020 \cite{Eurostat_2024}. Both of these statistics suggest that the construction industry will be facing major labor shortages in the future. In addition, the real labor productivity per hour worked has decreased compared to 2015 \cite{Eurostat_2024}. This particular trend does not apply to all industries, as the industrial sector without construction experienced an increase in real labor productivity \cite{Eurostat_2024}. A major difference between other industries and construction is the lack of adoption of digitization and automation in the construction industry. Construction companies see big potential in the adoption of these technologies, yet seem to have little or no implementation of these in place, according to a 2016 study conducted by Roland Berger \cite{schober_digitization_2016}. Another study focused on the United States construction industry, conducted by McKinsey\&Company, confirms this. In the study, construction is shown to be one of the sectors with the lowest adoption rate of digital technologies \cite{manyika_digital_2015}. Furthermore, it shows a correlation between the adoption rate of digital technologies and productivity growth, where construction is attested with the most negative growth rate between 2005 and 2015 \cite{manyika_digital_2015}.
\par
Another particularly detrimental fact is that construction consistently shows the highest incidence rate for work-related non-fatal accidents with 3,151.9 accidents per 100,000 people employed in 2021 \cite{Eurostat_2022_nonfatalaccidents} and the second-highest incidence rate for work-related fatal accidents with 6.32 deaths per 1000 people employed in 2021 \cite{Eurostat_2022_fatalaccidents}. Both issues, labor shortage, and workers' safety, can be alleviated by adopting digital technologies, especially developing autonomous robotic systems for construction activities. These systems would replace workers, addressing the labor shortage situation, and taking workers out of dangerous activities.
\par
One of the key challenges faced during the development and deployment of autonomous robotic systems in construction is the nature of the environment the systems need to operate in. For example, manufacturing environments, where robotic systems are widely adopted nowadays, are structured environments with clear, designated spaces for people, objects, etc. In addition, a manufacturing process's outcome is usually known in advance, making it easier to program robots to follow specific, pre-planned activities. This is generally not the case in construction, and this discrepancy is shown in \cite{parascho_construction_2023}.

To counterbalance the unstructured and highly dynamic nature of construction sites, au\-to\-no\-mous robotic systems need to rely on advanced perception capabilities, integrating sophisticated systems of sensors and algorithms to e.g., detect irregularities and navigate safely through a construction site. Most often, these detection algorithms are based on neural networks and require abundant data to train on. However, unlike other industries, the construction industry lacks this abundance of data. 
Our work addresses this issue and provides a new dataset focused on reinforced concrete construction processes, specifically the inspection and repair of reinforcement bars. The dataset consists of 14,805 RGB images, both from our own capturing sessions and publicly available data from YouTube. The images show different scenarios of shotcrete construction in both construction and repair phases, such as the construction of ground support walls, tunnels, and culverts, as well as the repair of piles and beams. 
With these images, we tackle the task of image segmentation for robotic inspection, repair, and construction of reinforced concrete structures. The images are provided with labels, which can be used to train deep learning-based segmentation algorithms for solving the segmentation task, which is a crucial element in enabling robots to autonomously perform construction tasks where reinforced concrete structures are involved. Such autonomous systems are currently in development, for example by the RobétArmé project \cite{kostavelis_robetarme_2024}, which is developing an autonomous robotic solution to shotcrete construction and benefits from datasets like ours for training deep learning-based methods for semantic understanding of the construction environment \cite{schmidt_towards_2024}. We demonstrate that the dataset can be used to train a working segmentation model by establishing baselines for the state-of-the-art models DeepLabV3, U-Net and YOLOv8L-seg. We perform experiments analyzing the influence of data availability on the performance of YOLOv8L-seg. We show that the model can be successfully used on an embedded GPU typically found in robotic platforms. Furthermore, as opposed to classic Computer Vision (CV) datasets, our dataset is scene-based and of sequential nature, i.e., the images have a temporal order, reflecting the spatio-temporal nature of robotic tasks. This property is used to identify label inconsistencies and analyze the effect of these on the performance of the model. Based on these experiments, we identify construction-related challenges that hinder the application of these technologies in the construction domain and propose steps to alleviate them.
\par
To summarize, the contributions of our work are as follows: 
\begin{itemize}
    \item A new open source dataset of construction imagery for reinforcement concrete construction with segmentation labels for four categories: exposed reinforcement bars, persons, cars, and trucks.
    \item An analysis protocol for label inconsistencies to analyze the influence of multiple valid labelling styles on model performance, i.e., opinions of different engineers regarding the extent of a defect.
    \item Introducing emphasis on spatio-temporal properties when employing CV techniques in a construction robotics context
    \item Investigating the influence of data availability and pre-training on the performance of the YOLOv8L-seg segmentation model.
    \item A discussion about challenges faced when applying deep learning-based methods in the construction sector, based on our experimental findings.
\end{itemize}

\section{Related Work}
The adoption and development of digital technologies in the Architecture, Construction, and Engineering (ACE) sectors is an ongoing and active field of research. The usage of autonomous systems relies on trustworthy and reliable perception capabilities of robots, which nowadays mostly rely on deep learning-based CV. 
\subsection{General overview}
CV can be employed for various tasks, starting from general-purpose site understanding tasks for robot navigation \cite{asadi_lnsnet_2019}, site monitoring for safety and management purposes \cite{yan_construction_2023, casuat_deep-hart_2020, bhadeshiya_hard-hat_2021, shan_du_hard_2011} to more specific process-related tasks: Numerous works cover the task of crack segmentation \cite{zhu_visual_2011, huthwohl_multi-classifier_2019, huang_nha12d_2022, tang2023a}, where traditional and deep learning-based computer vision techniques and models are deployed to do a pixel-level (dense) detection of cracks on surfaces, mostly concrete surfaces. A broader, associated task to this is the detection of damaged, reinforced concrete, where the goal is to detect not only surface cracks but also spallations, efflorescences, and exposed reinforcement bars (rebars) \cite{huthwohl_multi-classifier_2019, mundt_meta-learning_2019}. 
\subsection{Reinforcement bar detection}
\label{subsec:rebar_related_work}
Rebars are an important element at all lifecycle stages of structures, as they are of vital importance for their integrity. In early lifecycle stages, i.e., when structures are to be newly built, CV can be used to locate and assess the quality of rebar lattices/mats \cite{yuan_automatic_2023, chang_autonomous_2024, kardovskyi_artificial_2021, chen_intelligent_2023}, specifically the spacing and linkage between rebar elements as a crucial parameter to their contribution to the structural integrity of bridges and buildings. While lattices have a characteristic regular structure, there also exists work on locating individual bars, e.g., in precast reinforced concrete \cite{wang_automated_2017}. During the lifetime of a building or other infrastructure, it is important to monitor structural health. Defects can either appear on the surface in the form of spallations and exposed rebars---as mentioned previously, or underneath the surface. Sub-surface structural health can be surveyed using CV techniques on Ground Penetrating Radar (GPR) data \cite{liu_detection_2020, krause_image_2007, liu_using_2023} as opposed to commonly used RGB cameras and 3D Light Detection and Ranging (LiDAR) sensors. 
\subsection{Construction robots}
The previously mentioned related work focuses on inspection of structures and site monitoring and shows plenty of research activity in these areas. They provide a good basis for the adoption of autonomous systems, which rely on semantic understanding capabilities to navigate construction sites and carry out construction activities. Examples of such systems are excavators capable of autonomously building stone walls \cite{johns_autonomous_2020} or autonomous surface excavation \cite{jud_high-accuracy_2021}. Other examples are autonomous rebar tying robots \cite{cao_bim-based_2024, zhang_efficient_2024} and autonomous 3D concrete printing robots with rebar placement \cite{du_bim-enabled_2024}. A pipeline detecting exposed rebars from RGB images and reconstructing a 3D model from stereo images for autonomous robotic placement of shotcrete is presented in \cite{schmidt_towards_2024, robotics13070102}. \cite{taha_robotic_2019} developed a proof of concept showcasing the robotic application of shotcrete, and \cite{frangez_depth-camera-based_2021} shows an algorithm to identify and reconstruct steel rebar meshes. A thorough review of the state of construction robotics can be found in \cite{xiao_recent_2022}.
\subsection{Datasets}
The lack of publicly available datasets in the ACE domain is a common problem and poses a barrier to the development of deep learning-based solutions. The few (semi-)publicly available datasets addressing the ACE domain range from general-purpose object detection datasets for site monitoring and navigation to more task-specific datasets. SODA \cite{duan_soda_2022} is such a general-purpose dataset consisting of around 20,000 images with around 268,200 bounding boxes, depicting 15 classes of categories \textit{Person}, \textit{Material}, \textit{Machine} and \textit{Layout}. In \cite{duan_soda_2022}, other general-purpose datasets are listed as well, such as MOCS \cite{xuehui_dataset_2021}, a dataset for detecting moving machinery and workers, ACID \cite{xiao_development_2021} for detecting construction machinery, and CHV \cite{wang_fast_2021} for detecting personal protective equipment (PPE). CIS \cite{yan_construction_2023} is another general-purpose dataset for detecting people, machines, and materials. However, as opposed to the aforementioned datasets, this dataset features instance segmentation annotations instead of just bounding boxes. Out of these five datasets, only CHV and CIS are publicly available, while the others are available only upon request.
\par
Task-specific datasets are more tailored to solving a specific construction task. Table \ref{tab:rebar_datasets} gives an overview of datasets used in the works in Section \ref{subsec:rebar_related_work} depicting rebars for various tasks. As the table shows, availability of data is low, which poses a barrier to the development of robust, deep learning-based solutions. Not every work makes a data availability statement, and more often than not, the data is only available upon request. Only \cite{wang_synthetic_2023} provides their data without any additional barriers. We want to overcome this barrier by releasing our dataset to the public, hoping that it will motivate other researchers to make also their data publicly available.

\renewcommand{\arraystretch}{.75}
\begin{table}
    \centering
    \caption{Examples of rebar detection datasets}
    \begin{tabular}{l>{\raggedleft\arraybackslash}p{.2\textwidth}>   {\raggedleft\arraybackslash}p{.2\textwidth}>{\raggedleft\arraybackslash}p{.2\textwidth}}
        \hline
         &\cite{jin_robotic_2021}& \cite{liu_detection_2020} & \cite{kardovskyi_artificial_2021}\\
         \hline
         No. images& 256 & 3992 & 240 \\
         No. objects& more than 10,000 crosspoints & 13,026 & N/A \\
         Modality & RGB-D & Ground penetrating radar & RGB + Stereo\\
         Annotation type & Keypoints & Bounding Boxes & Instance Segmentation \\
         Image sizes & N/A & N/A & 1280$\times$720\\
         Domain & real & real & real \\
         Task & Crosspoint detection for rebar tying & Non-invasive localization of rebars in concrete & Quality assessment of rebar installations \\
         Availability & No availability statement & No availability statement & No availability statement \\
         \hline
    \end{tabular}\\
    \vspace{.5cm}
    \begin{tabular}{l>{\raggedleft\arraybackslash}p{.2\textwidth}>   {\raggedleft\arraybackslash}p{.2\textwidth}>{\raggedleft\arraybackslash}p{.2\textwidth}}
        \hline
         & \cite{wang_synthetic_2023} & \cite{chang_autonomous_2024} & \cite{chen_intelligent_2023} \\
         \hline
         No. images& 2500 & 192 & 3130 \\
         No. objects& N/A & N/A & N/A \\
         Modality & RGB & RGB + LiDAR & RGB \\
         Annotation type & Instance Segmentation & N/A & Semantic Segmentation \\
         Image sizes & 1280$\times$720 & N/A & N/A \\
         Domain & synthetic & real & real \\
         Task & Instance Segmentation of Rebars & Instance Segmentation of Rebars for spacing analysis & Segmentation of Rebars for Quality Control \\
         Availability & public & Upon request & Upon request \\
         \hline
    \end{tabular}
    \label{tab:rebar_datasets}
\end{table}

\section{The ConRebSeg Dataset}
The following section introduces and carries out a deep analysis of our dataset. We will start with general properties such as image sizes, number of objects, and object sizes, and will also analyze aspects that are more specific to our dataset. 
\subsection{Overview}
We have collected a dataset of 14,805 images captured mostly in construction environments. The main purpose of this dataset is to collect a broad set of images depicting exposed rebars in the context of shotcrete construction, aimed at autonomous inspection, repair, and construction of reinforced concrete structures. Our dataset contains scenes from various shotcrete construction tasks, such as the construction of ground support walls, tunnels, and culverts as well as the repair of piles and beams.  This captures the diverse appearance of exposed rebars. The scenes consist of temporally ordered images, reflecting the robotic application this dataset targets.\par
We provide segmentation masks for four different classes:
\begin{itemize}
    \item \textbf{Exposed rebars:} These can be whole rebar lattices, i.e., in ground support wall construction, or partially exposed lattices and single bars in defective concrete.
    \item \textbf{Persons:} These will usually be construction workers in PPE.
    \item \textbf{Cars:} Those are regular-sized passenger cars.
    \item \textbf{Trucks:} This is usually heavy construction machinery or material delivery trucks.
\end{itemize}
Given the embedding of this work in the RobétArmé project \footnote{\url{https://www.robetarme-project.eu/}}, which develops an autonomous robotic system for shotcrete inspection, repair and construction, we put our main focus on the exposed rebars class since it is important to identify these areas as they need to be filled with shotcrete. The other classes were chosen as sanity check classes, aimed at supervising the annotation process and empirically assessing the quality of the annotations.\par
In total, there are 54,115 instances, distributed as follows: 41,237 exposed rebars, 11,275 persons, 330 cars and 1273 trucks. Our dataset is a mix of self-collected data, for which the setup is described in Section \ref{sec:acquision_setup}, and YouTube videos.

\subsection{Acquisition of data and labeling process}
\label{sec:acquision_setup}
The scenes in the dataset stem from two sources: self-collected data we acquired during construction site visits and data from YouTube. The camera we used in our data collection visits is a Basler a2A1920-160ucBAS RGB camera fitted with a Basler C125-0418-5M-P f4mm lens. The camera's exposure time is set to 5 ms, with automatic white balancing and the light source preset is set to \texttt{Tungsten2800K}. To avoid noisy images, we set the noise reduction and sharpness enhancement parameters to -10,000. We performed two data collection sessions for our self-captured data. The collection has been conducted in a hand-held manner, so as not to introduce any platform-specific characteristics into the data, possibly jeopardizing a general applicability of the dataset on different robotic platforms. A summary of the two recording sessions is included in the first two columns of Table \ref{tab:self_collected_sequences}.

\renewcommand{\arraystretch}{1.5}
\begin{table}[htbp]
    \centering
    \caption{Metadata of the three constituent parts of ConRebSeg}
    \begin{tabular}
    {>{\raggedright\arraybackslash}p{.25\textwidth}>{\raggedleft\arraybackslash}p{.2\textwidth}>   {\raggedleft\arraybackslash}p{.2\textwidth}>{\raggedleft\arraybackslash}p{.2\textwidth}}
        \hline
         Location &  Vester Sogade, 1601 Copenhagen, DK & Langebro, 1219 Copenhagen, DK &\textcolor{black}{Web (YouTube)}\\
         Short Description & Balcony repair works with exposed rebars, both individual and in lattice format & Repair of bridge piles and beams in challenging lighting conditions &\textcolor{black}{Mix of ground support walls, piles and beams, and tunnels and culverts}\\
         \textcolor{black}{Environmental conditions} & \textcolor{black}{outdoors, cloudy, naturally illuminated} & \textcolor{black}{indoors, dark, artificially illuminated} & \textcolor{black}{varying}\\
         Number of scenes & 12 & 14 &\textcolor{black}{51}\\
         Number of frames & 4911 & 7270 &\textcolor{black}{2624}\\
         Average frame rate & 1.8 FPS & 3.8 FPS  &\textcolor{black}{N/A}\\ 
         Resolution (width/height) & 1920x1200 & 1920x1200 & \textcolor{black}{varying}\\
         \hline
    \end{tabular}
    \label{tab:self_collected_sequences}
\end{table}

Arranging construction site visits always relies on partners with connections to the construction industry, which can be a great barrier in the process, since not every site owner is willing to let outsiders in. That is why in addition to the self-collected data, we have also identified publicly accessible videos on YouTube depicting shotcrete construction activity, which has allowed us to rapidly grow the dataset. However, the frames of those videos are not directly included in the dataset, but rather their annotations together with a mapping to a specific frame of the video. This is due to legal uncertainties regarding the distribution of downloaded YouTube videos \footnote{In their service terms, YouTube states that "\textit{You are not allowed to:} [\ldots]\textit{1. access, reproduce, download, distribute, transmit, broadcast, display, sell, license, alter, modify or otherwise use any part of the Service or any Content except: } \textit{[\ldots] (c) as permitted by applicable law;}" and "[\ldots] \textit{access the Service using any automated means (such as robots, botnets, or scrapers) except}: [\ldots] \textit{(c) as permitted by applicable law;}" \cite{youtube_terms_nodate}. We stipulate that the EU DSM directive \cite{noauthor_directive_2019} and its national implementation in Denmark \cite{noauthor_elov_nodate} are fulfilling the conditions on which users are allowed to download data from YouTube as per their service terms.}. Instead, we provide a way for users to obtain the data themselves in an automated manner for scientific/research purposes. Users should check the legality of this process in their respective jurisdictions.
Once we have collected the data, we annotated them to obtain masks for the previously listed object classes. To make the annotation process privacy-preserving, we have treated all scenes with the face-blurring algorithm Deface \cite{noauthor_orb-hddeface_2024}, based on the CenterFace \cite{xu_centerface_2020} face detector. Furthermore, we have treated all frames of the scenes gathered at Langebro (see Table \ref{tab:self_collected_sequences}) with histogram equalization and AI-denoising based on NAFNet \cite{chen_simple_2022} to alleviate the rather low-illumination, low-contrast frames. This facilitates the annotation process and avoids low-quality annotations due to invisible image features. However, we provide the original, but anonymized images in our dataset to give users the freedom to pre-process the frames according to their needs while maintaining the needed level of privacy. The experiments conducted in Section \ref{sec:experiments} use the raw version of the images without any level of anonymization, and we stipulate that the effect on the outcome of the experiments remains minimal.
\par
\begin{figure}[htbp!]
    \centering
    \begin{subfigure}{.3\textwidth}
        \includegraphics[width=\textwidth]{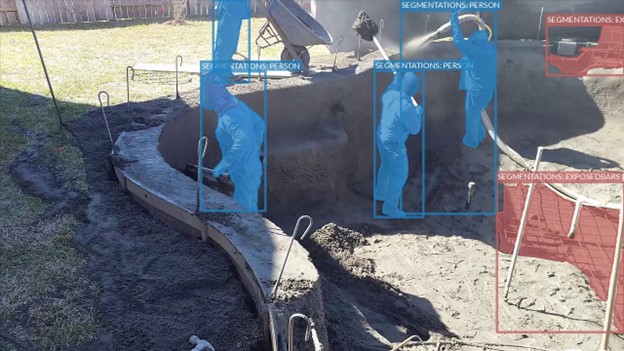}
    \end{subfigure}
    \begin{subfigure}{.3\textwidth}
        \includegraphics[width=\textwidth]{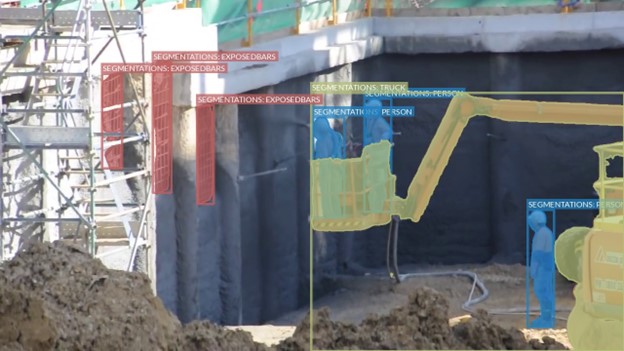}
    \end{subfigure}
   \begin{subfigure}{.3\textwidth}
        \includegraphics[width=\textwidth]{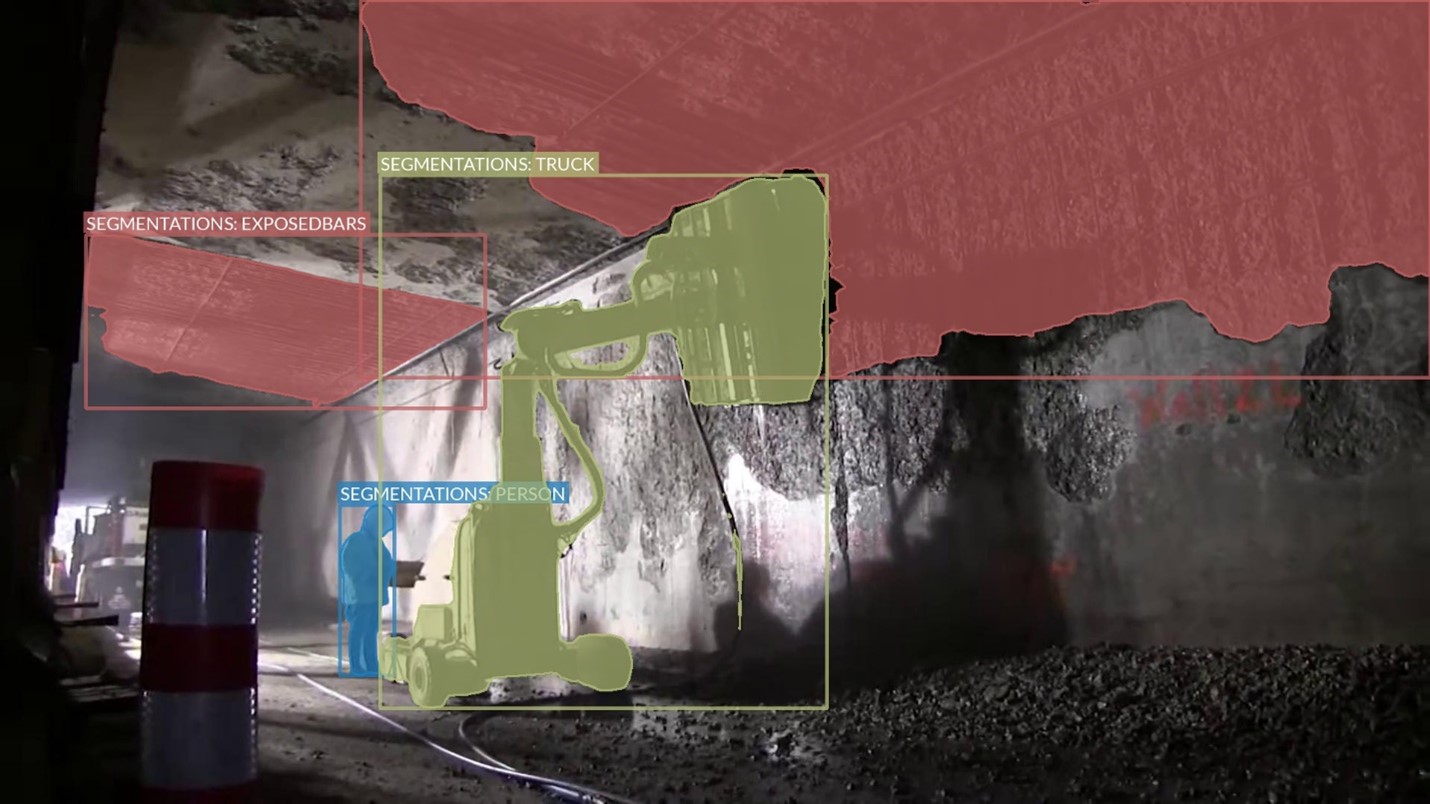}
    \end{subfigure}
   \begin{subfigure}{.3\textwidth}
        \includegraphics[width=\textwidth]{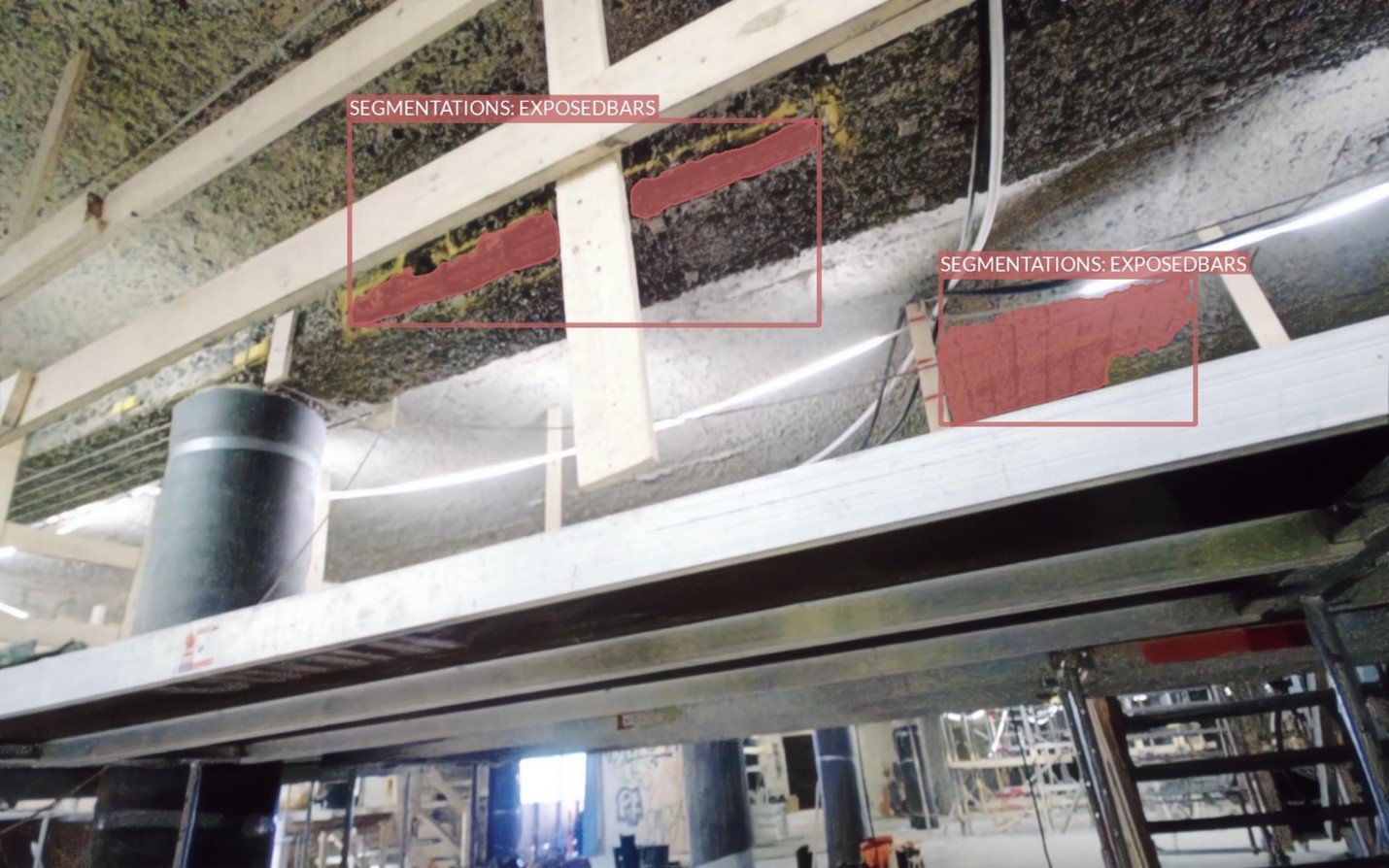}
    \end{subfigure} 
   \begin{subfigure}{.3\textwidth}
        \includegraphics[width=\textwidth]{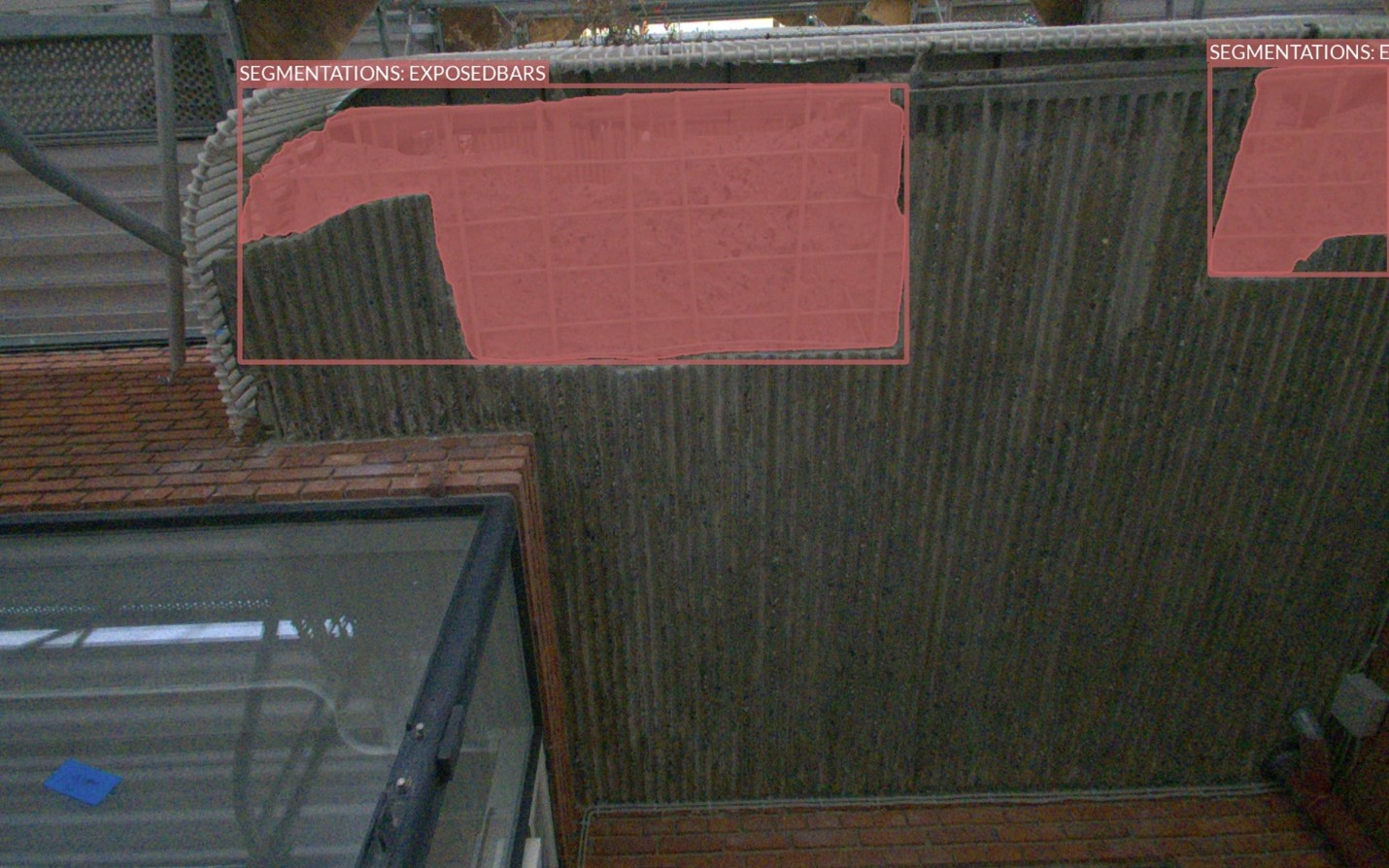}
    \end{subfigure} 
   \begin{subfigure}{.3\textwidth}
        \includegraphics[width=\textwidth]{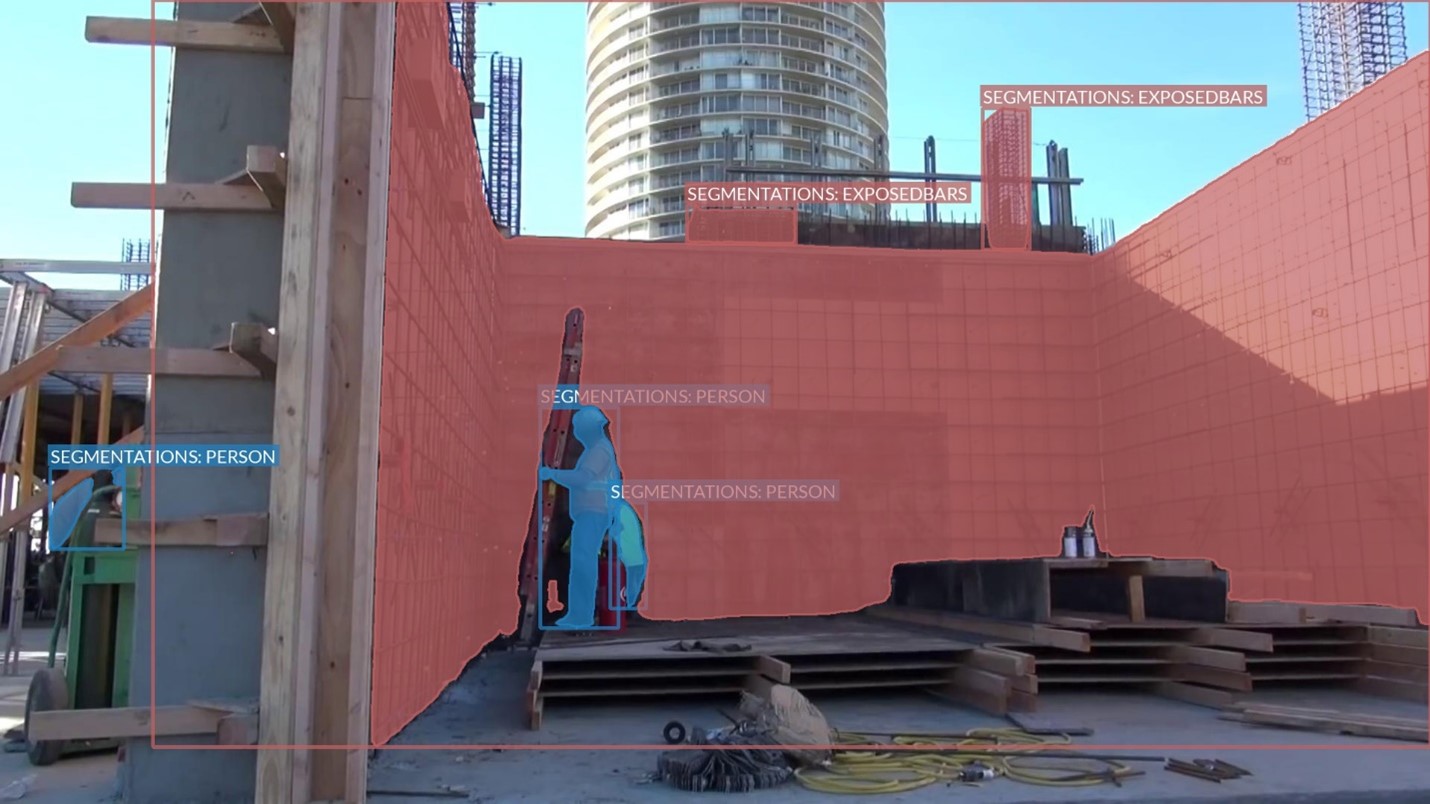}
    \end{subfigure} 
    \caption{Visual examples provided as annotation guidelines}
    \label{fig:annot_examples}
\end{figure}

While classes such as \textit{person}, \textit{truck}, and \textit{car} have some common sense associated with them, the \textit{ExposedBars} class is rather domain-specific and not easy to describe. We set the following description for the annotation process: \textbf{area} with exposed rebars, where the tips of the bars should be polygon points and roughly outline the area. Along that, we gave some visual examples, as illustrated in Figure \ref{fig:annot_examples}.
\subsection{Statistics}
\begin{figure}[htbp]
    \centering
\includegraphics[width=0.7\textwidth]{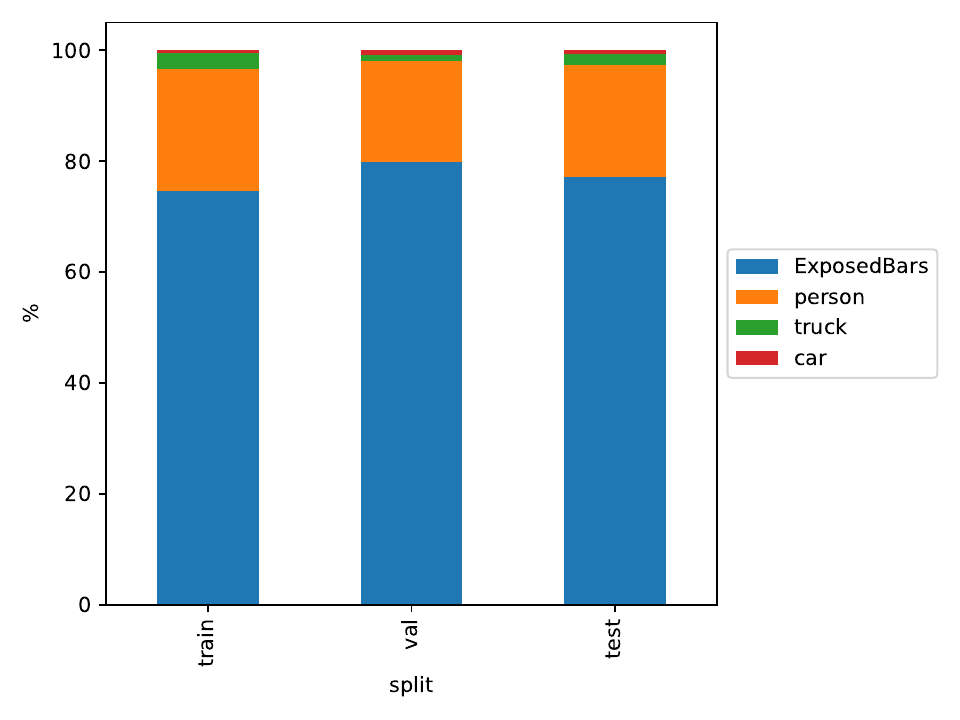}
    \caption{Class distribution within each split}
    \label{fig:dataset_split}
\end{figure}

Our dataset is split into a training (train), validation (val), and testing (test) set. The split was done ensuring an equal distribution of class instance counts. Figure \ref{fig:dataset_split} shows the relative distribution of the count of object instances in the respective splits.
Figure \ref{fig:dataset_split} also shows a clear dominance of the \textit{ExposedBars} class, followed by the \textit{person} class. We neglect this class imbalance as we are mainly interested in the ExposedBars category, given that there exists a plenitude of segmentation models trained on general-purpose datasets like MS COCO \cite{coco_dataset}, Pascal VOC \cite{pascal-voc-2012} or Cityscapes \cite{cordts_cityscapes_2016} for the remaining categories in our dataset.
In total, there are 8570, 3543, and 2692 samples for the train, val, and test split respectively. Table \ref{tab:label_stats} shows key statistics about the splits and their annotations. 

\begin{table}[htbp]
    \centering
\caption{Label statistics, per dataset split}
\label{tab:label_stats}
    \begin{tabular}{lrrr}
        \hline
         &  train&  val& test\\ 
         \hline
         Average number of objects per image&  3.79&  3.01& 4.08\\ 
         Average mask size in pixels&  77,151&  47,408& 115,337\\ 
         Average bounding box size in pixels&  141,703&  133,770& 200,729\\
         \hline
    \end{tabular}
\end{table}

Table \ref{tab:label_stats} shows that the average object count per image stays roughly the same throughout the splits, however, the mask and bounding box size vary. The train set is a balance point, as it lies between the val and test split in terms of object count and size.
\begin{figure}[htbp]
    \centering
    \includegraphics[width=\textwidth]{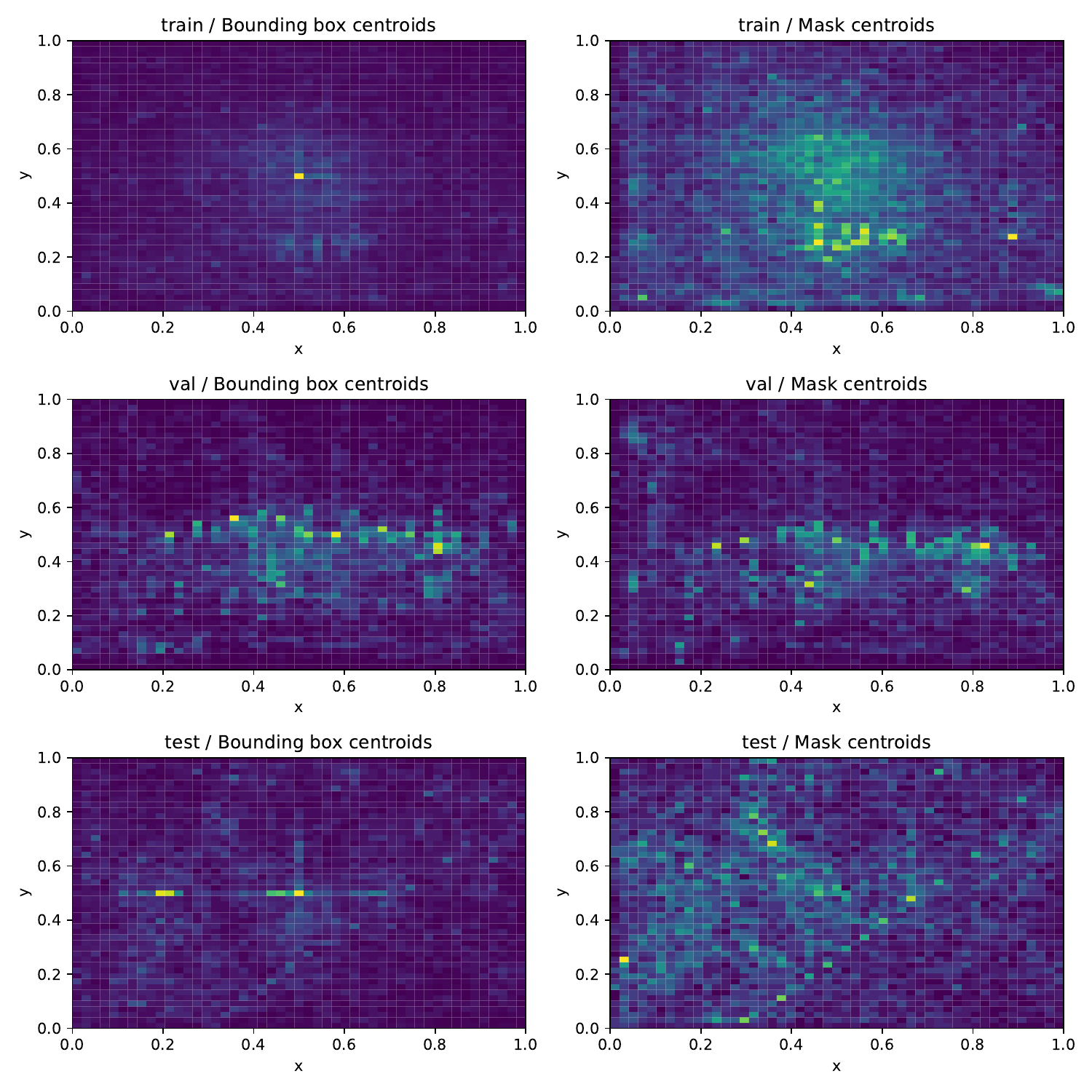}
    \caption{Histogram of normalized label centroids in the train (upper row), val (middle row), and test split (bottom row) of the ConRebSeg dataset, for the mask bounding boxes (left) and for the mask itself (right)}
    \label{fig:label_centroids}
\end{figure}
Figure \ref{fig:label_centroids} shows histograms of the centroids of bounding boxes on the left and masks on the right respectively, for the train, val, and test split in the first, second, and last row respectively. Especially for the train split we can observe a clear bias towards the center of the picture, which isn't as expressed in the val and test splits, where the centroids lie still vertically centered, however with some horizontal spread.

\subsection{Distribution and structure of the dataset}
\begin{figure}[htbp]
    \begin{forest}
      for tree={%
        folder,
        grow'=0,
        fit=band,
        font=\tt
      }
      [data/
        [langebro/
            [run\_<year>\_<month>\_<day>\_<hh>\_<mm>\_<ss>/
                [\_<year>\_<month>\_<day>\_<hh>\_<mm>\_<ss>.<dddd>.jpg]
            ]
        ]
        [vester\_sogade/
            [\_<year>-<month>-<day>-<hh>-<mm>-<ss>/
                [<ns\_epoch\_timestamp>.png]
            ]
        ]
        [youtube/
            [youtube\_id/
                [frame\_<\%04d>.jpg]
            ]    
        ]
      ]
    \end{forest}
    \caption{File system structure of the dataset}
    \label{fig:filesystem_structure}
\end{figure}
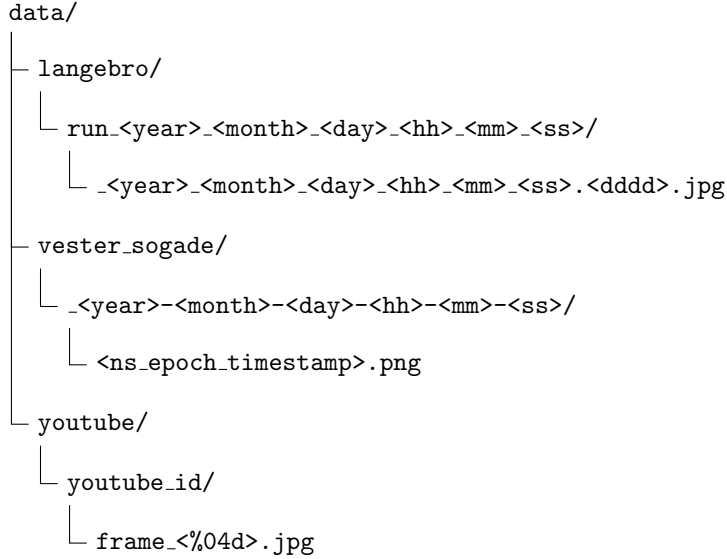

We provide a public repository that contains our data and other elements that facilitate viewing, exploring, and experimenting with the dataset. We use the FiftyOne dataset management framework \cite{noauthor_fiftyone_nodate} that provides end-users with a web interface for viewing samples and managing metadata such as train/test/val tags as well as other tags describing certain properties of samples. Our repository contains scripts that set up the dataset for the user, especially obtaining the data from YouTube, as we do not provide the frames directly for reasons specified in Section \ref{sec:acquision_setup}. In general, the data is organized as listed in Figure \ref{fig:filesystem_structure}.

The angle brackets in Figure \ref{fig:filesystem_structure} are placeholders and denote the meaning of the fields. \texttt{dddd} are decimal seconds, \texttt{ns\_epoch\_timestamp} denotes a timestamp in nanoseconds in Unix Epoch time (since Jan. 1, 1970), \texttt{\%04d} means that frame numbers up to 1000 are zero-padded.
\par
The YouTube frames are downloaded as whole videos and then are converted to frames, however, only every 200$^{th}$ frame is used since the informational value of the frames in between is minimal due to minimal variations of the depicted scene. For integrity purposes, we save a MD5 checksum for every sample in the dataset, which is saved as a metadata field in the dataset management tool. At import time, every frame is checked against this checksum to ensure the integrity of the dataset.

\subsection{Annotation Styles / Cleanup}
\label{subsec:annotation_styles}
\begin{figure}[htbp]
    \centering
    \begin{subfigure}{.3\textwidth}
        \includegraphics[width=\textwidth]{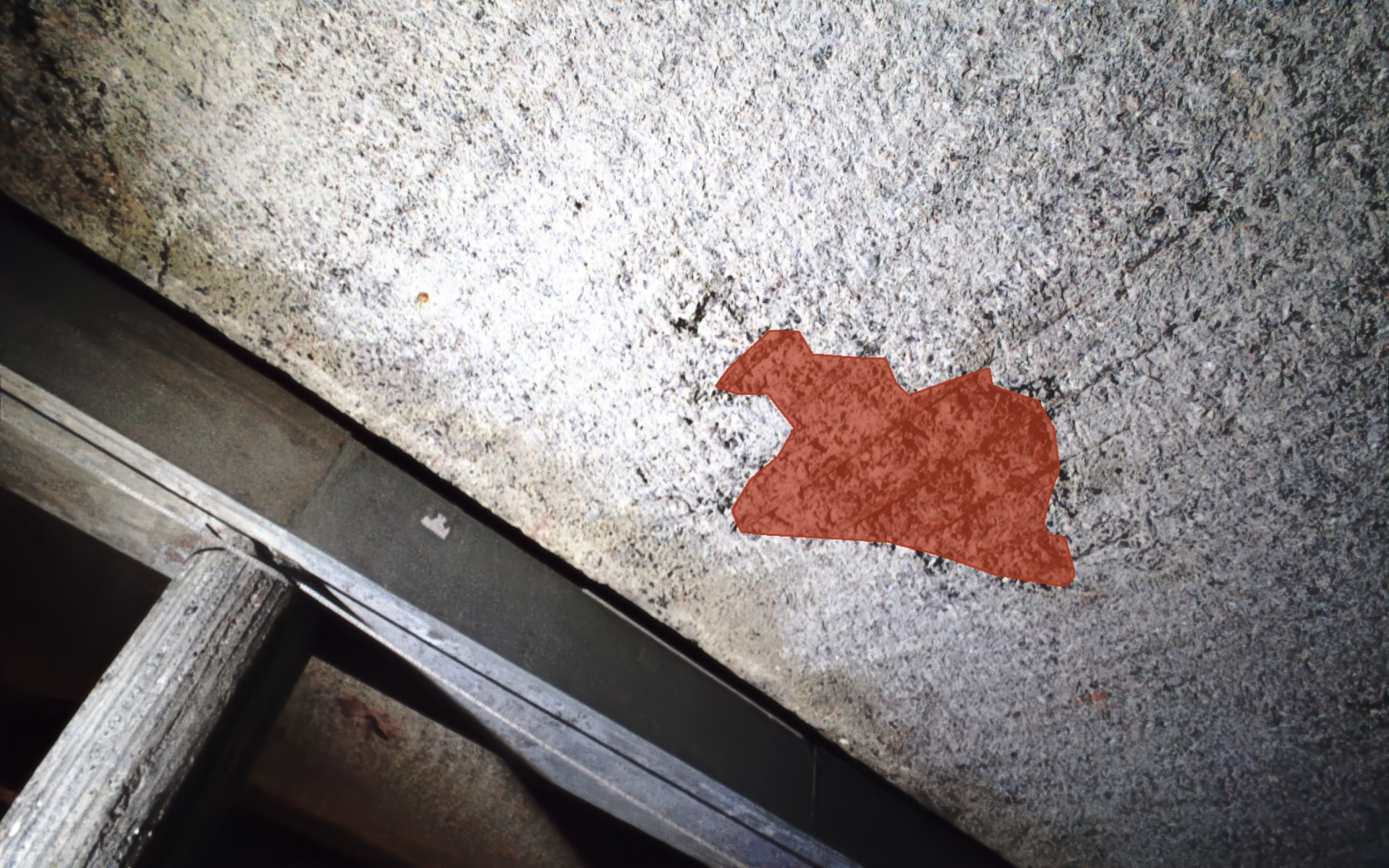}
    \end{subfigure}
    \begin{subfigure}{.3\textwidth}
        \includegraphics[width=\textwidth]{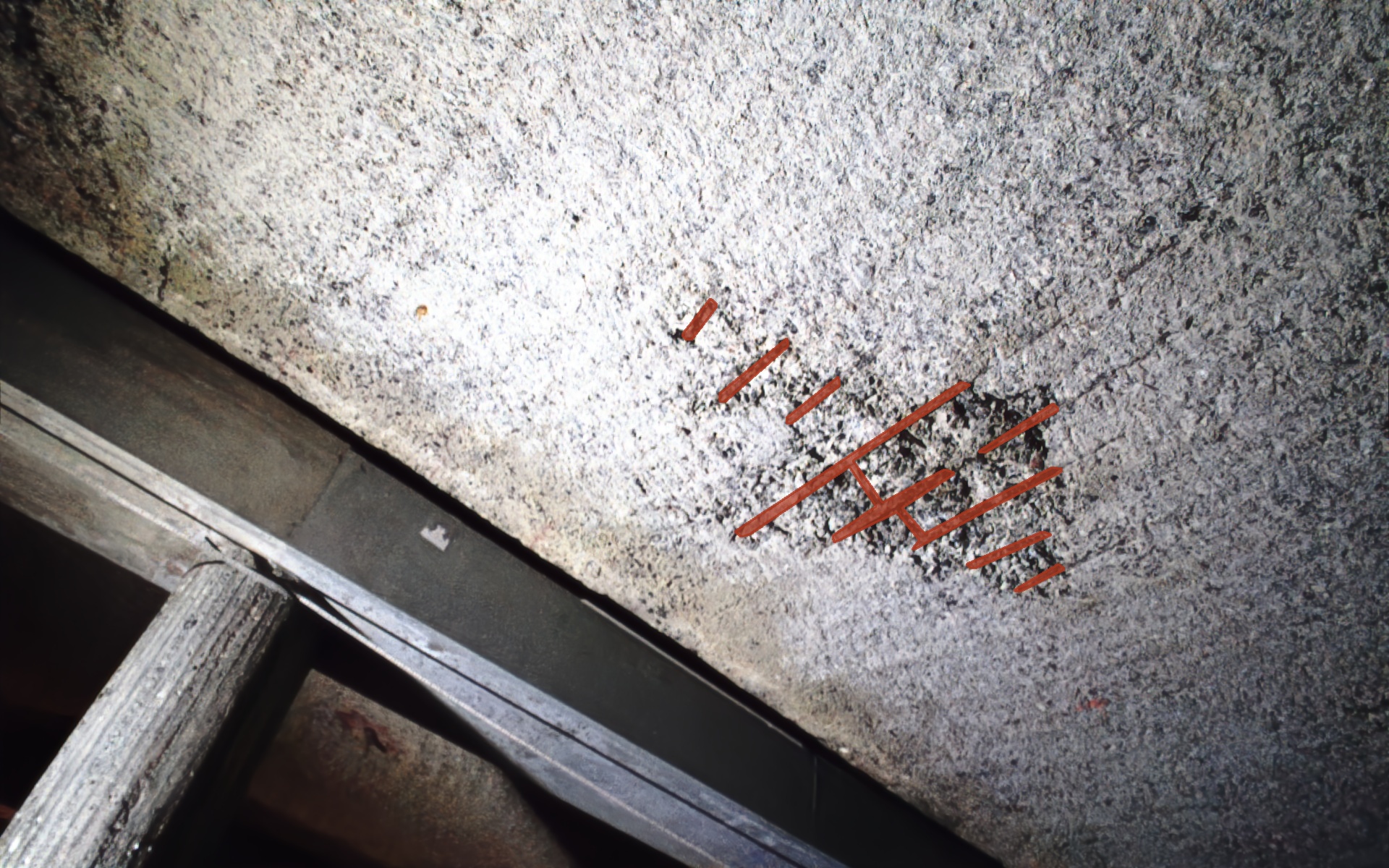}
    \end{subfigure}
       \begin{subfigure}{.3\textwidth}
        \includegraphics[width=\textwidth]{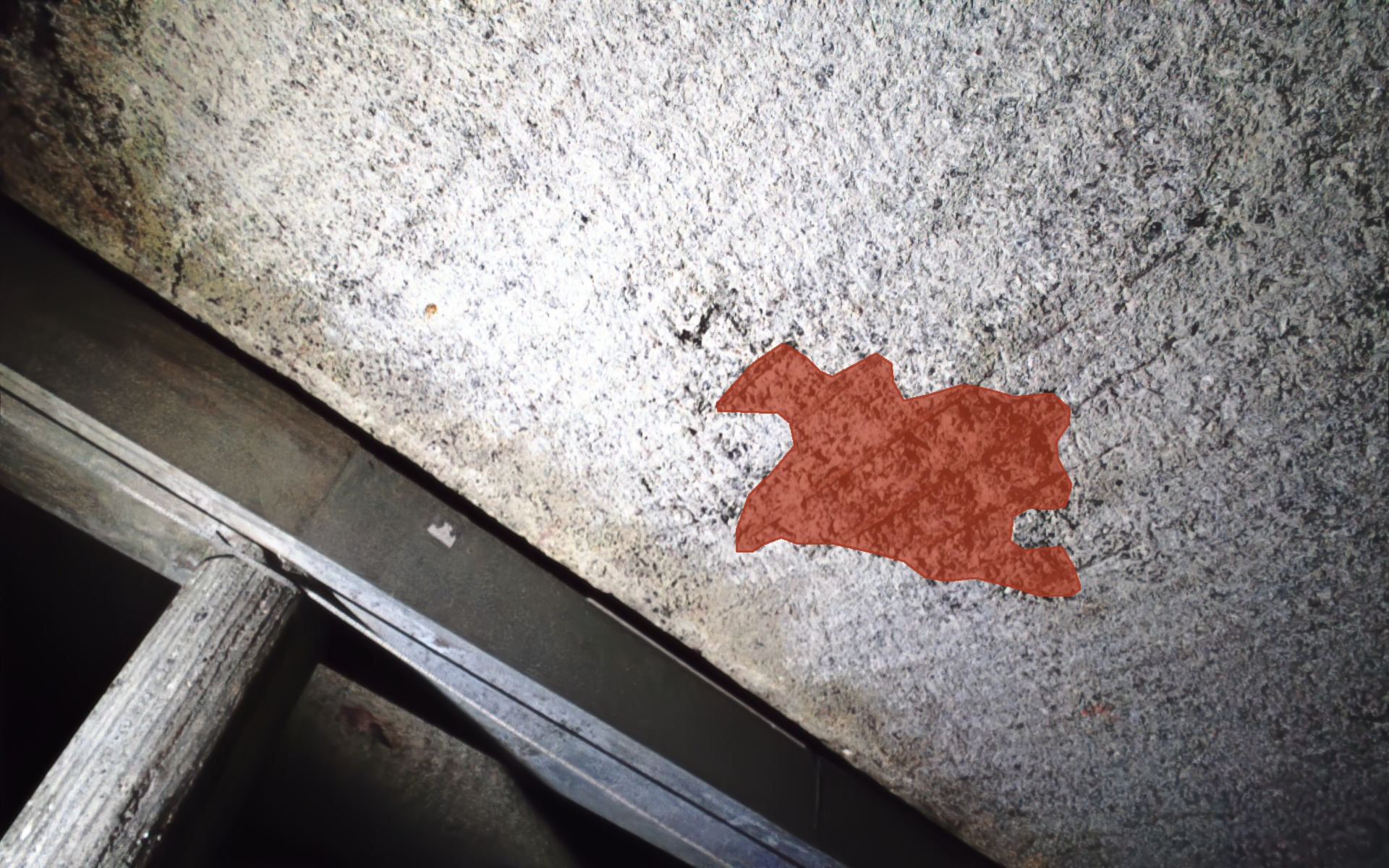}
    \end{subfigure}
    \caption{Sequence of successive frames with their labels (in red) showing an inconsistency in labeling style. The left and right frame share the same annotation style, the center frame has a different style.}
    \label{fig:annotation_styles}
\end{figure}

A phenomenon we observed while analyzing the segmentation masks is that the same object has been annotated in different ways. This was to be expected since we shuffled the frames within scenes collected at Langebro (see Table \ref{tab:self_collected_sequences}) to remove the temporal dependencies within frames since we wanted to analyze how different annotators will label the \textit{ExposedBars} category. Figure \ref{fig:annotation_styles} shows three consequent frames with minimal changes depicting the same object, along with the mask.

It can be seen that the mask drastically changes in appearance, moving from a continuous "area" annotation to a more fine-grained, "rebar" annotation. While our label description explained that "areas" should be annotated, we conclude that both styles are valid annotations that serve different purposes. The styles could reflect "expert" opinions, which both need to be respected. However, we assume that two vastly different masks for very similar images might impede the training of deep learning-based segmentation models. We will elaborate on this in Section \ref{subsec:train_without_anomaly} in more detail, and describe an approach to separate the different styles in the following.
\par
\begin{figure}[htbp]
    \centering
        \includegraphics[width=\textwidth]{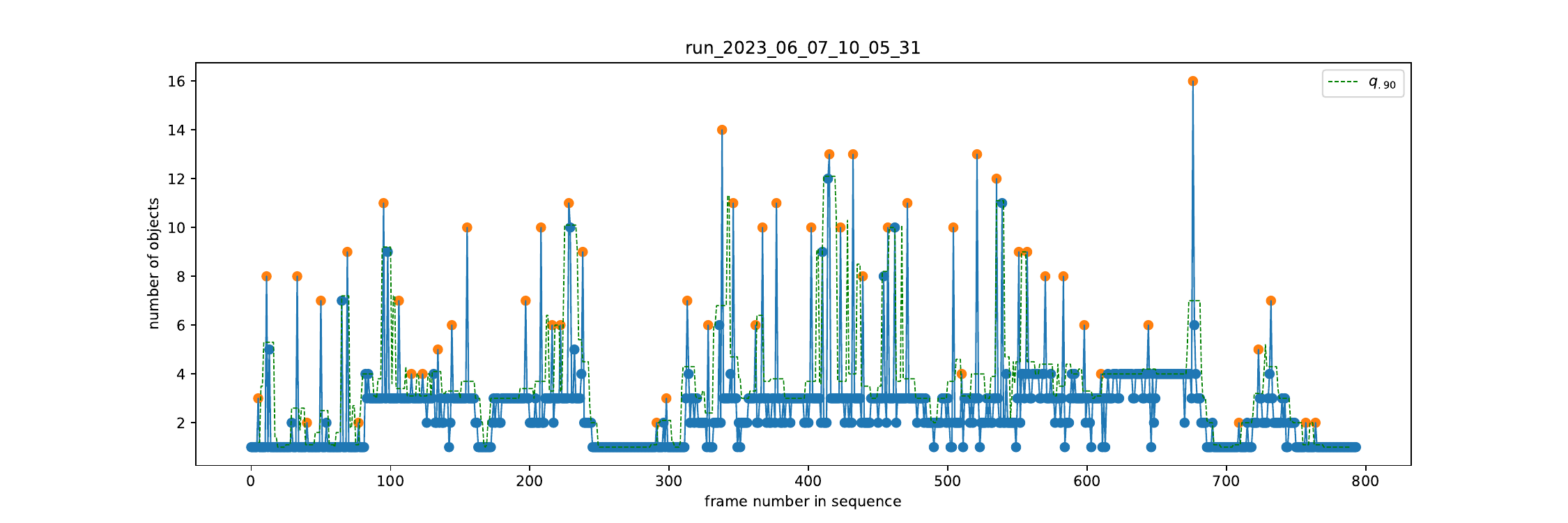}
    \caption{Plot of the number of objects $n_{obj}(t)$ (on y-axis) per frame (on x-axis) in the Langebro scene \texttt{run\_2023\_06\_07\_10\_05\_31}. Orange dots indicate anomalies. The dashed green line indicates the local .90 quantile $q_{.90}(t)$}
    \label{fig:num_objects_langebro_10_05_31}
\end{figure}

Figure \ref{fig:annotation_styles} suggests an indicator being the temporal development of the number of objects (per-class) in an image. The object count from the left picture in Figure \ref{fig:annotation_styles} to the middle one suddenly increases from 1 to 7 and then goes back to 1 in the last picture. We, therefore, extract the number of unconnected masks of the same class per frame and create a time series from it, respecting the temporal order of the frames. A local inconsistency is then determined by a rolling window approach of taking $n=10$ frames centered around the current frame, determining their 0.90 quantiles $q_{.90}(t)$ of the object count, and marking a frame as an anomaly if the number of objects in the frame exceeds $q_{.90}(t)$. Formally, the rule is expressed as follows:
\begin{equation}
    \text{anomaly}(t)=
    \begin{cases}
        \text{1}, &\text{if } n_{\text{obj}}(t) > q_{.90}(t) \\
        \text{0}, & \text{otherwise}
        \end{cases}
\end{equation}
with 1 indicating an anomaly and 0 indicating that something is adhering to the "common" style. We now use the train scene \texttt{run\_2023\_06\_07\_10\_05\_31} from the Langebro scenes as an example. Figure \ref{fig:num_objects_langebro_10_05_31} shows a plot of the time series, with the number of objects $n_{obj}(t)$ in a frame noted on the y-axis. The green dashed line indicates $q_{.90}(t)$, and any point above this line is considered an anomaly and is denoted as an orange dot.  Based on this rule, we tag the frames within a scene with an \textit{anomaly} tag if they fulfill it.

\begin{figure}[htbp]
    \centering
    \hfill
    \begin{subfigure}{.49\textwidth}
            \includegraphics[width=.49\textwidth]{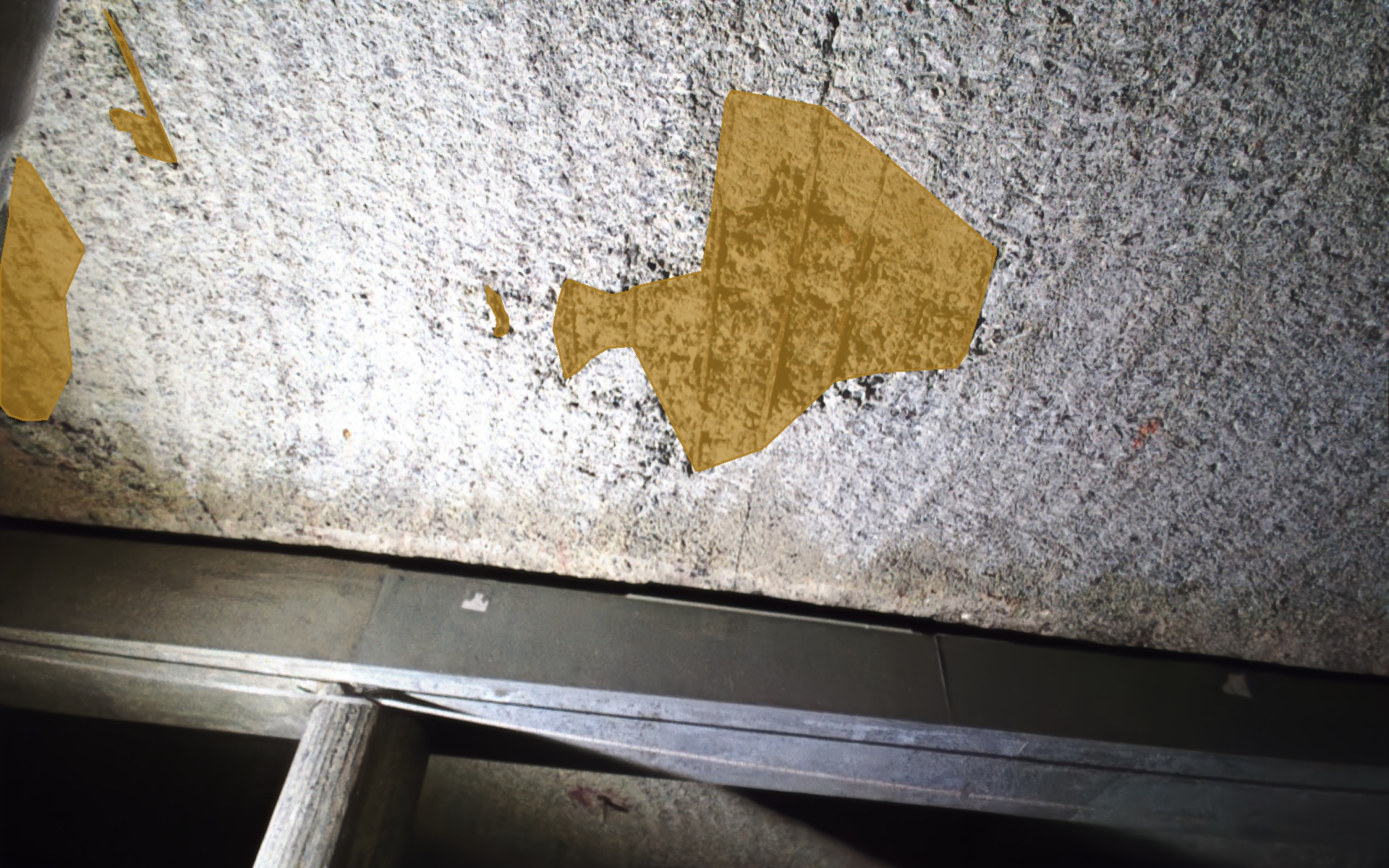}
            \includegraphics[width=.49\textwidth]{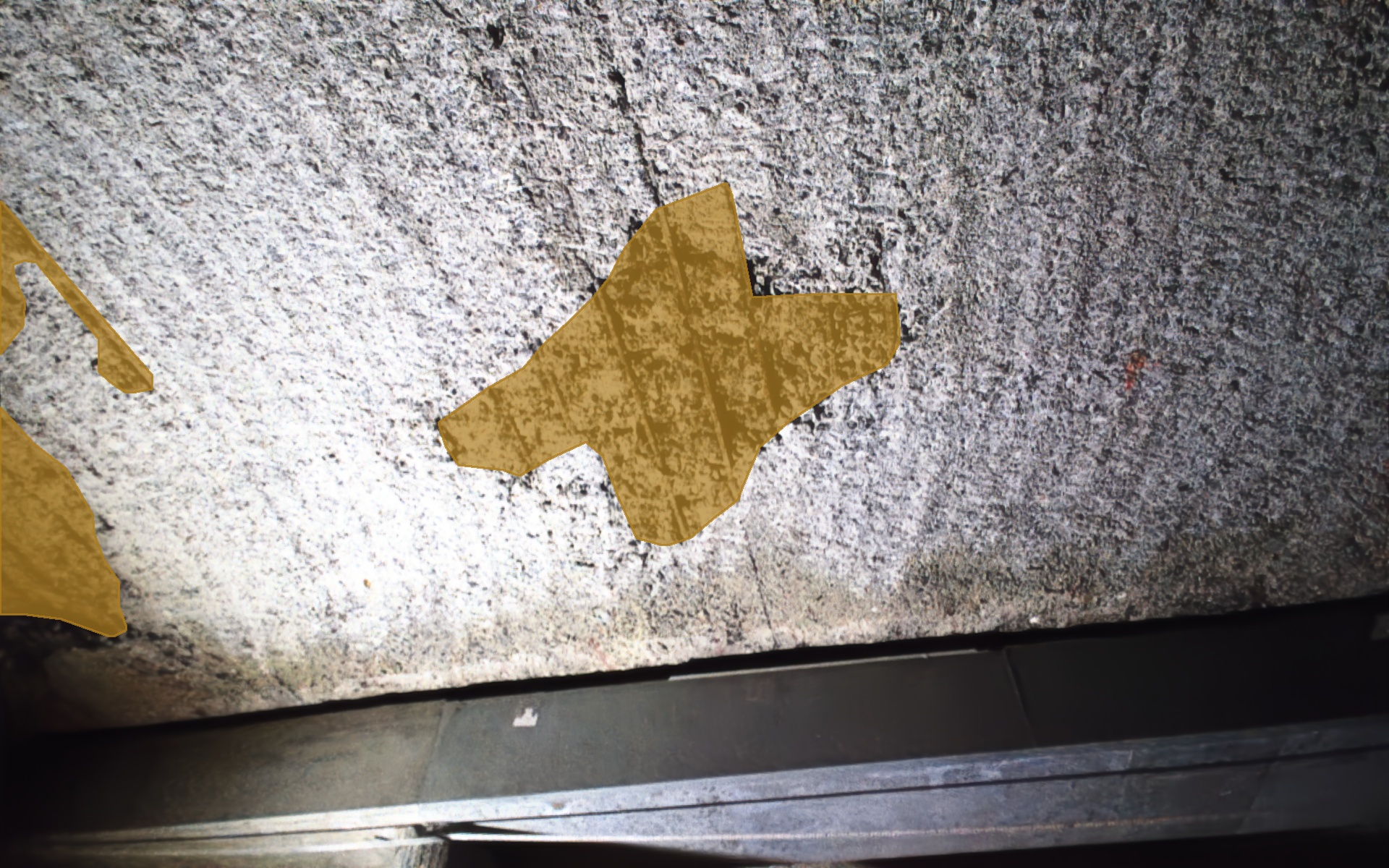}\\
            \includegraphics[width=.49\textwidth]{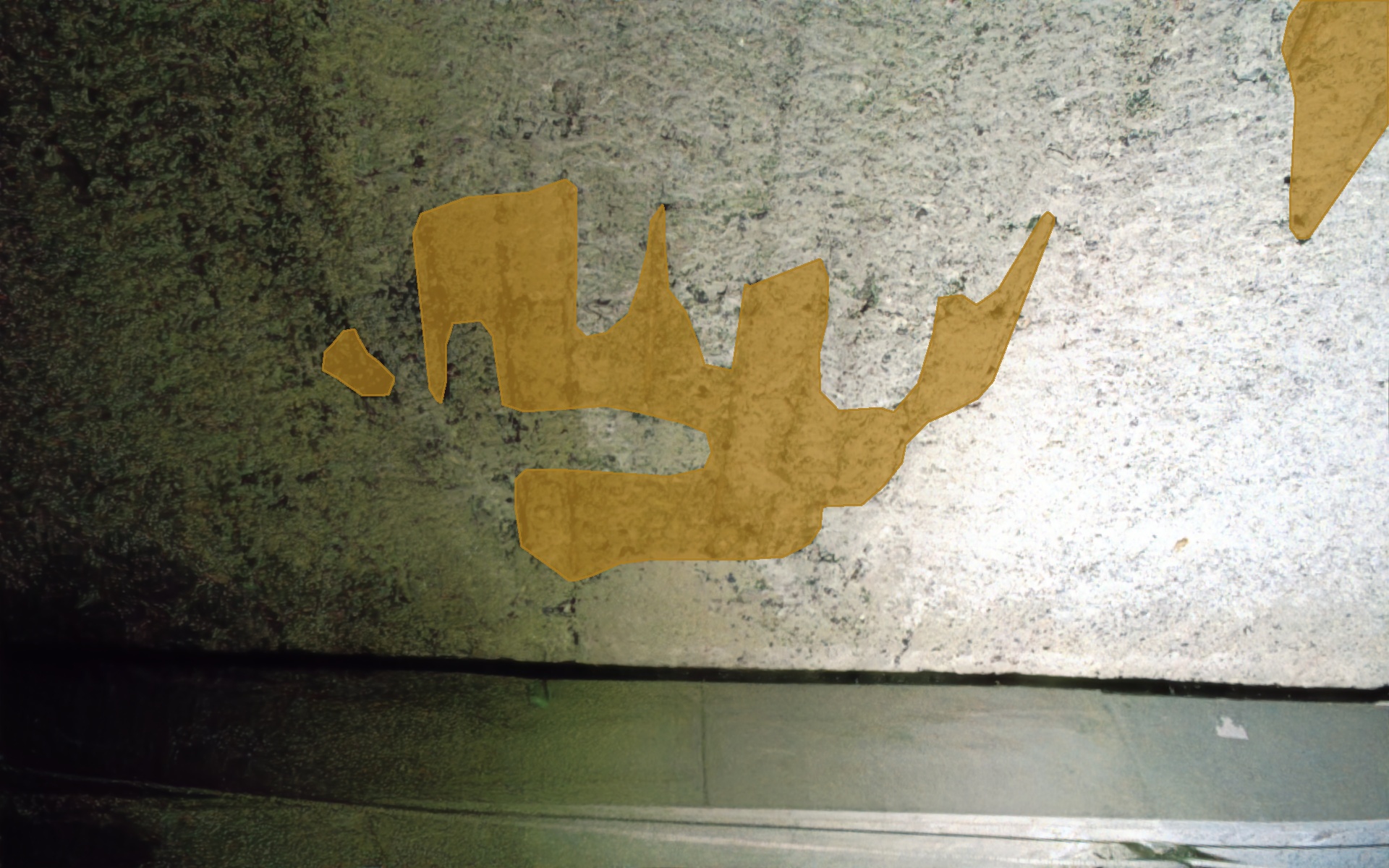}
            \includegraphics[width=.49\textwidth]{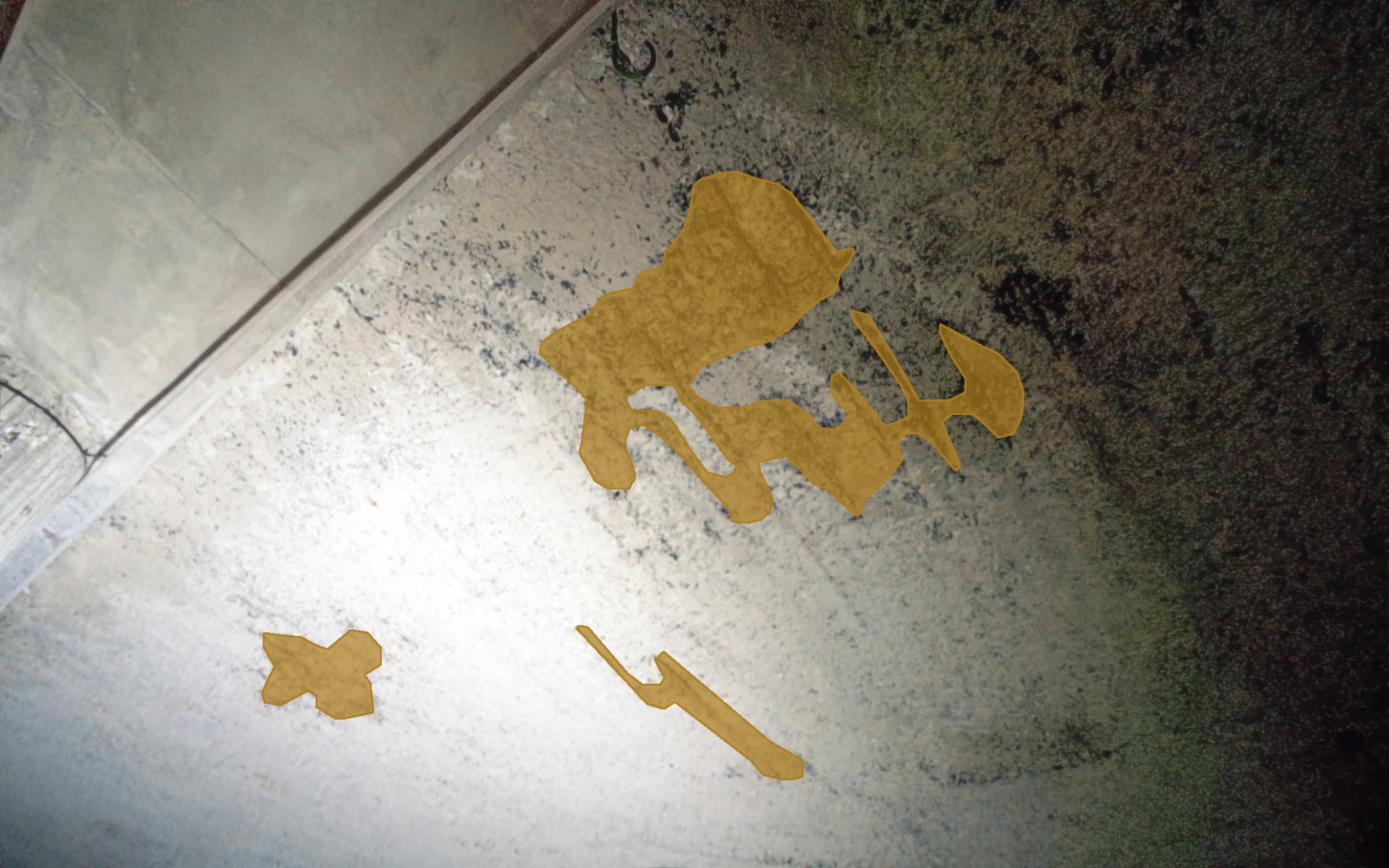}
        \caption{Randomly sampled non-anomaly frames}
        \label{fig:non_anomaly_10_05_31}
    \end{subfigure}
    \hfill
    \begin{subfigure}{.49\textwidth}
       \includegraphics[width=.49\textwidth]{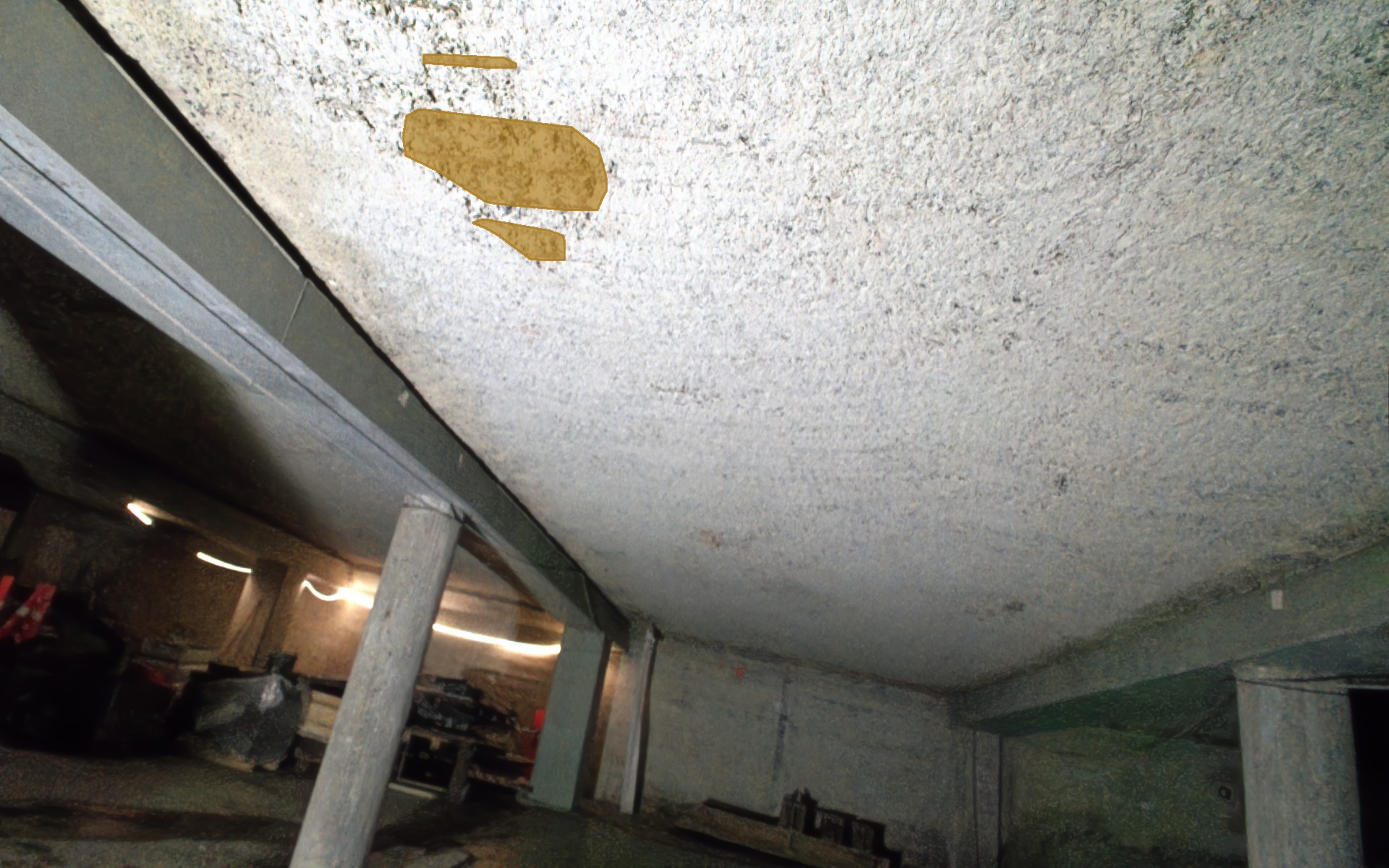}
       \includegraphics[width=.49\textwidth]{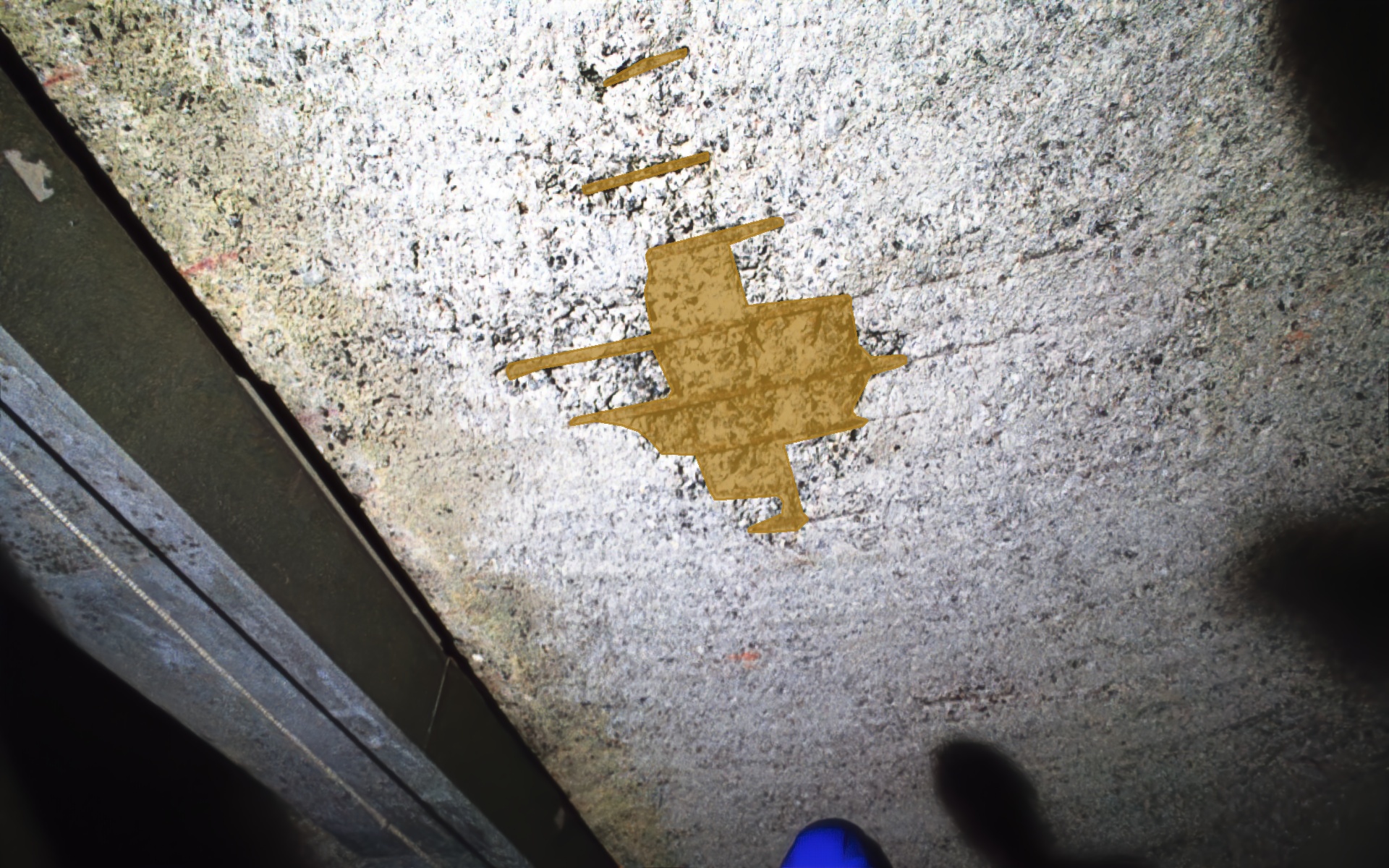}\\
       \includegraphics[width=.49\textwidth]{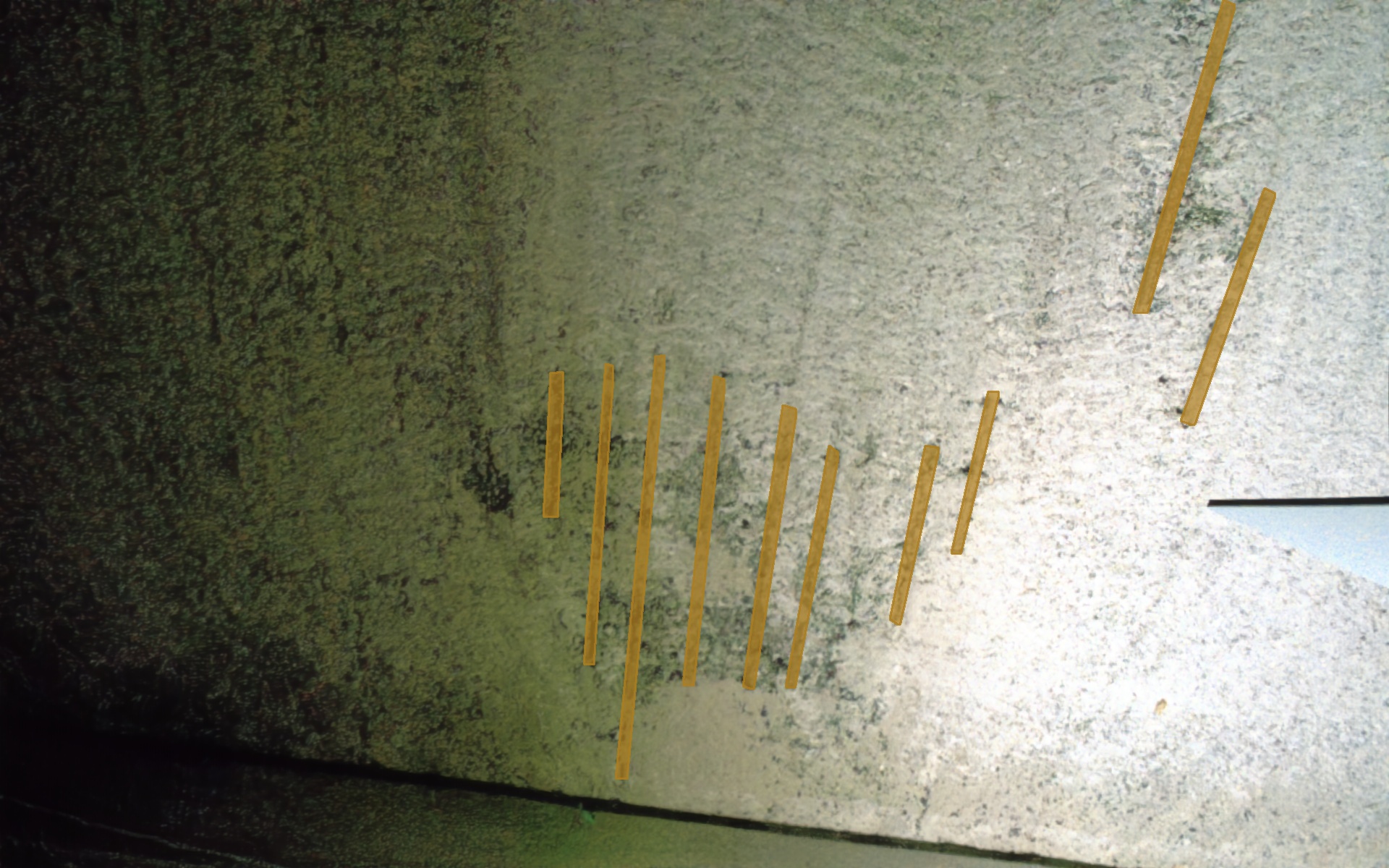}
       \includegraphics[width=.49\textwidth]{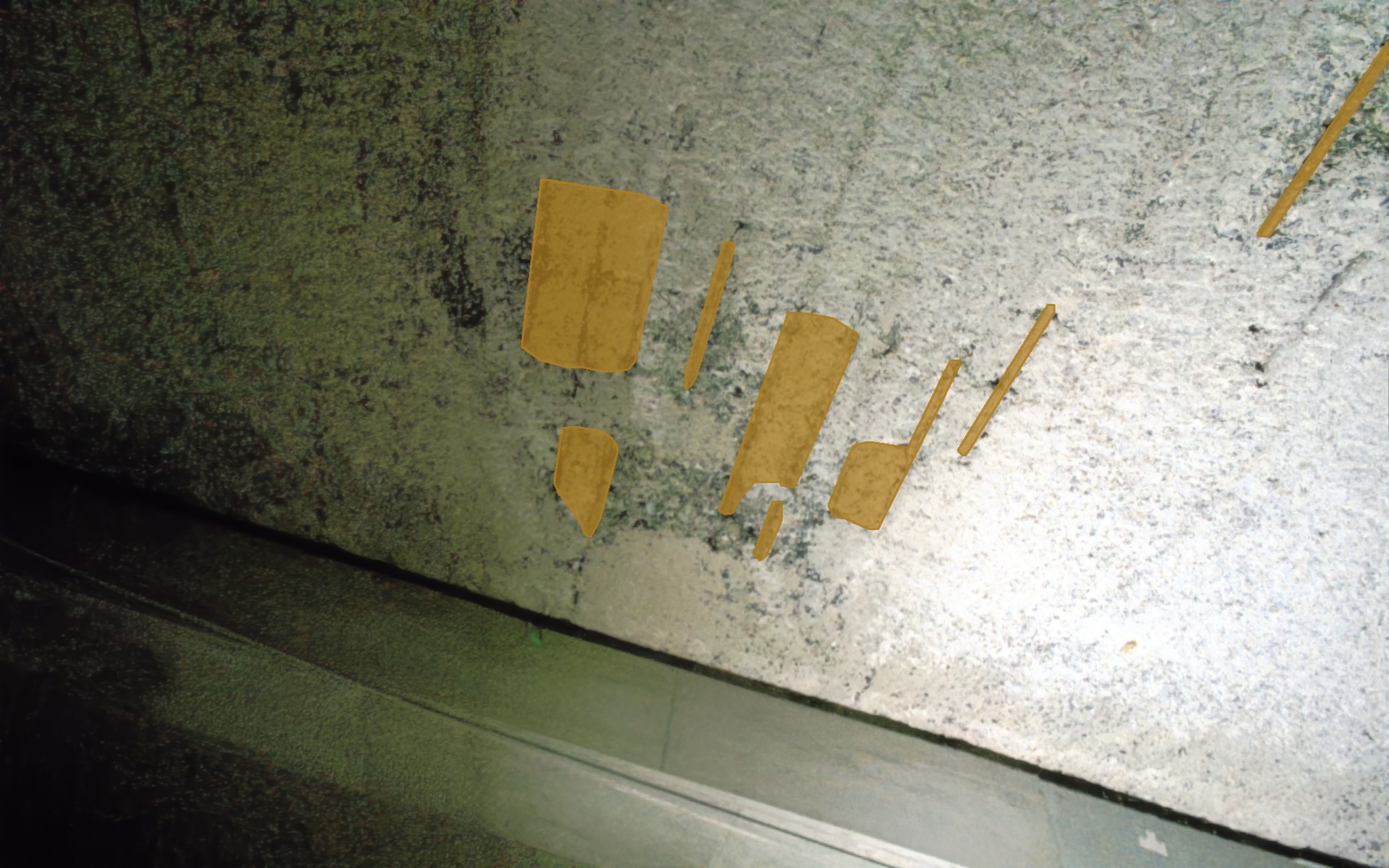}
       \caption{Randomly sampled anomaly frames}
       \label{fig:anomaly_10_05_31}
    \end{subfigure}
    \hfill
    \vspace{1em}
    \caption{Comparison between frames and their labels (in yellow) without the anomaly tag (left) and with the anomaly tag (right)}
\end{figure}

Figures \ref{fig:non_anomaly_10_05_31} and \ref{fig:anomaly_10_05_31} show the result of the tagging process on Langebro scene \\\texttt{run\_2023\_06\_07\_10\_05\_31}. We have randomly sampled four frames from samples without (Figure \ref{fig:non_anomaly_10_05_31}) and with the anomaly (Figure \ref{fig:anomaly_10_05_31}) tag. Clearly,  the samples tagged as anomalies show the more fine-grained, single-bar annotation style as opposed to an area annotation. We therefore proceed to use the tags in Section \ref{subsec:train_without_anomaly} to distinguish between the labels and carry out experiments related to this phenomenon.
\section{Experiments}
\label{sec:experiments}
This section presents the various experiments conducted on the dataset. We will first present our infrastructure used in the experiments. Then, we will establish some baselines for different semantic and instance segmentation models,  and carry out experiments regarding data efficiency and the influence of certain labelling styles on the performance of the models.
\subsection{Setup}
For our experiments, we use three different setups, as listed in Table \ref{tab:hw_setup_experiments}. All the models used in the experiments are based on PyTorch \cite{Ansel_PyTorch_2_Faster_2024} and trained on CUDA-enabled GPUs. To ensure consistent and comparable results within and across models, we enable deterministic training and initialize the random number generators with fixed seeds. 

\begin{table}
    \centering
\caption{Hardware Setups used in our experiments}
\label{tab:hw_setup_experiments}
    \begin{tabular}{l>{\raggedleft\arraybackslash}p{0.25\linewidth}>{\raggedleft\arraybackslash}p{0.25\linewidth}>{\raggedleft\arraybackslash}p{0.25\linewidth}}
        \hline
          &A&  B& C\\
          \hline
          CPU&Intel Xeon Gold \newline 6126 / 6142 / 6242&  Intel Xeon Gold \newline 6226R / 6326& AMD Ryzen Threadripper PRO 5945WX\\
          GPU&NVIDIA Tesla V100&  NVIDIA Tesla A100& NVIDIA GeForce RTX 3090\\
 OS& Scientific Linux 7.9& Scientific Linux 7.9&Ubuntu 22.04 LTS\\
        \hline
    \end{tabular}

\end{table}
\subsection{Baseline Training}
This section establishes baselines for various models to compare future experiments against. We will establish baselines for 3 models: YOLOv8L-seg, DeepLabV3, and U-Net. U-Net and DeepLabV3 are semantic segmentation models, i.e., produce a dense pixel map for the whole image. YOLOv8L-seg is an instance segmentation model, i.e., it is based on an object detection model and produces a binary mask within each bounding box detection. The aforementioned models are all CNN-based models. 
\par
Regarding the handling of the output of instance segmentation models, we assemble an image-level mask by pooling the instance predictions and their corresponding masks together, i.e., we perform a logical OR operation. If two detections occupy the same pixel and have different class labels, we resolve this by assigning the pixel the class label of the prediction with the higher confidence score.  Unless otherwise specified, we train each model with the recommended default hyperparameters, and only keep detections and their corresponding masks with a confidence score equal to or greater than 0.25, and report macro-level metrics. This means the metric is computed per category and then averaged with equal weighting. All models except U-Net have around 45 million trainable parameters, making it fair to compare them against each other.

In the first round, we train from scratch with default optimizer hyperparameters, and we only consider data from the first 100 epochs. For all models, we use an image size of 640$\times$640 pixels. YOLOv8L-seg uses extensive image augmentations, whereas DeepLabV3 and U-Net use only \texttt{RandomHorizontalFlip (RHF)} and \texttt{RandomVerticalFlip (RVF)} augmentations, both with $p=0.5$. Table \ref{tab:baseline_hyp} summarizes the hyperparameters used for each model.

\begin{table}[htbp]
    \centering
    \textcolor{black}{
    \begin{threeparttable}[htbp]
    \caption{Training hyperparameters used in establishing the baselines}
    \label{tab:baseline_hyp}
    \begin{tabular}{>{\raggedright\arraybackslash}p{0.2\linewidth}>{\raggedleft\arraybackslash}p{0.2\linewidth}>{\raggedleft\arraybackslash}p{0.2\linewidth}>{\raggedleft\arraybackslash}p{0.2\linewidth}}
    \hline
         &  DeepLabV3&  U-Net& YOLOv8L-seg\\
         \hline
         Image size&  640x640&  640x640& 640x640\\
         Batch size&  8&  8& 16\\
         Augmentations&  \texttt{RHF} and \texttt{RVF} with $p=0.5$&  \texttt{RHF} and \texttt{RVF} with $p=0.5$& recommended default \tnote{1}\\
 Weight Initialization& scratch& scratch&scratch\\
         Optimizer&  Adam with \texttt{PolynomialLR} scheduler&  Adam& SGD with momentum\\
         Optimizer learning rate&  0.001&  0.001& 0.01, momentum 0.9\\
 Epochs& 200& 200&200\\
    \hline
    \end{tabular}
    \begin{tablenotes}
        \item[1]{\url{https://docs.ultralytics.com/usage/cfg/?h=augmentation\#augmentation-settings}}
    \end{tablenotes}
    \end{threeparttable}
    }
\end{table}

\begin{table}
    \centering
    \caption{Baseline results}
    \begin{tabular}{lrrr}
        \hline
         &  DeepLabV3&  U-Net&  YOLOv8L-seg\\
         \hline
         Checkpoint Epoch&  44&  78&  100\\
         mIOU (train)&  0.705 &  0.574 &  0.687\\
 mIOU (val)& 0.341 & 0.301 & 0.516\\
 Mask mAP$_{50-95}$ (val)& N/A& N/A& 0.216 \\
 Box mAP$_{50-95}$ (val)& N/A& N/A& 0.282 \\
 \hline
    \end{tabular}
    \label{tab:baseline_scratch_results}
\end{table}
Table \ref{tab:baseline_scratch_results} shows the results of the baseline experiments. While DeepLabV3 and U-Net achieve similar performances in regard to mean Intersection Over Union (mIOU) on the val split (mIOU $\approx$ 0.3), YOLOv8L-seg almost performs a magnitude better with a val mIOU of approximately 0.5. Given the superior performance of YOLOv8L-seg compared to the other two models, we proceed to conduct further experiments based on the YOLOv8L-seg model. A straightforward experiment is to start the training from pre-trained weights provided by the developers of the model. Those pre-trained weights stem from pre-training on the large-scale MS COCO dataset, which consists of 330,000 images \cite{coco_dataset}. The result of this experiment is listed in Table \ref{tab:baseline_pretrained_results} and shows improvement compared to the model trained from scratch. Therefore, trainings we conduct in further experiments will also be started from pre-trained weights.
\begin{table}
    \centering
    \caption{Influence of pre-trained weights on model performance}
    \begin{tabular}{l>{\raggedleft\arraybackslash}p{0.3\textwidth}>{\raggedleft\arraybackslash}p{0.3\textwidth}}
    \hline
         &  YOLOv8L-seg (scratch) &YOLOv8L-seg (pre-trained)\\
         \hline
         Checkpoint Epoch&  100 &58\\
         mIOU (train)&  0.687&0.681\\
 mIOU (val)& 0.516 &0.544 \\
 Mask mAP$_{50-95}$ (val)& 0.216 
&0.317 \\
 Box mAP$_{50-95}$ (val)& 0.282 &0.374 \\
 \hline
    \end{tabular}
    \label{tab:baseline_pretrained_results}
\end{table}

\subsection{Effect of withheld training data}
\label{subsec:data_efficiency}
With the established baselines being trained on 100\% of the training data, we want to analyze the influence of withheld data on the performance of the models. This gives an intuition about how much variation the data introduces and how diverse our dataset is. There are two ways to withhold the data; one can either withhold whole training scenes, or withhold samples from every training scene. We opted for the former to actually reduce the variation in data.

\begin{figure}[htbp]
    \centering
    \begin{subfigure}{.49\textwidth}
        \includegraphics[width=\textwidth]{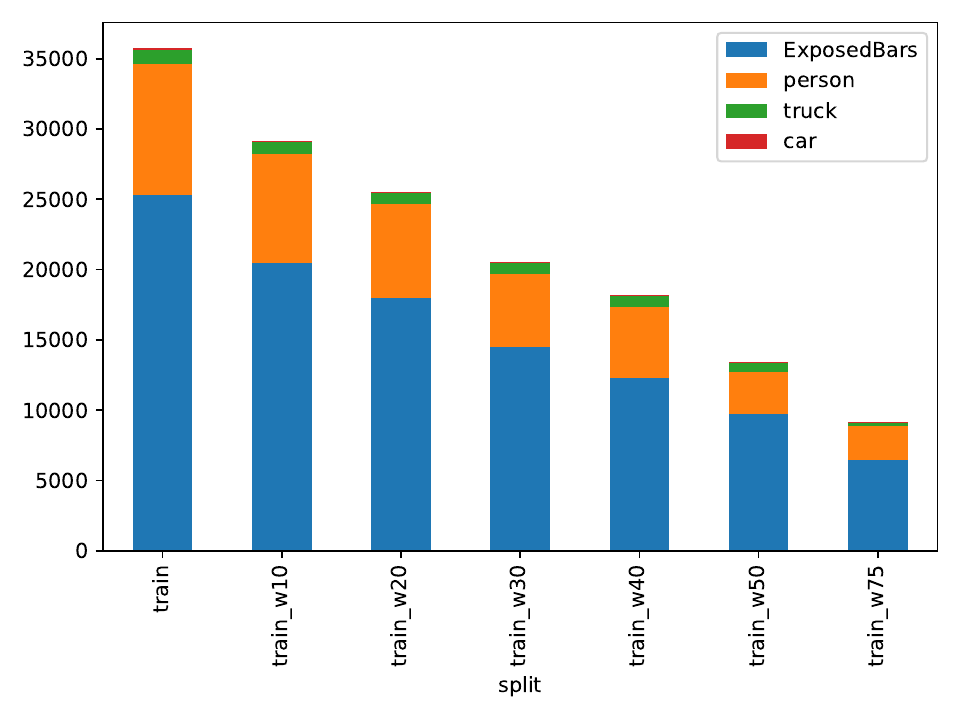}
    \end{subfigure}
    \begin{subfigure}{.49\textwidth}
        \includegraphics[width=\textwidth]{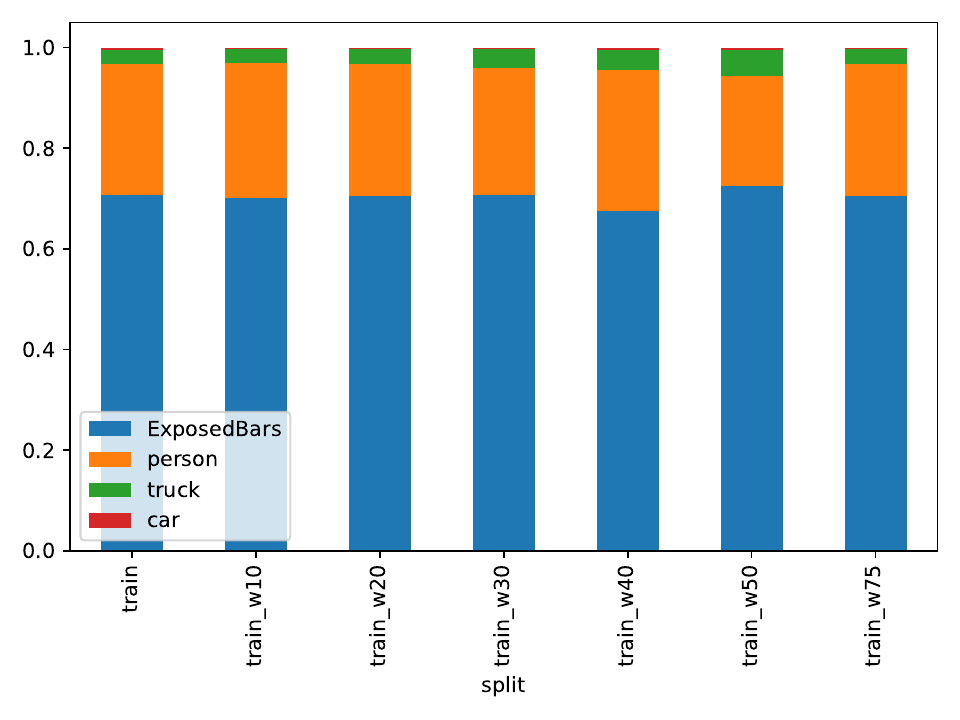}
    \end{subfigure}   
    \caption{Instance class distribution of train split, in absolute values (left) and relative values (right) for different amounts of withheld training data}
    \label{fig:withhold_split_dist}
\end{figure}

It is important to ensure that the withheld data doesn't change the data distribution too much, as this drift would introduce performance implications already. Furthermore, it makes comparing the various experimental results unfair. Figure \ref{fig:withhold_split_dist} shows the instance class distributions for each new split \texttt{train\_wX}, where \texttt{X} is the percentage of withheld training data. The percentage is calculated on the number of samples, not the number of instances. The left plot shows the decreasing number of instances with more and more withheld data, while the right plot shows the approximately equal class distribution, with the first bar showing the class distribution of the original train split without any withheld data. 
\par
\begin{figure}[htbp]
    \centering
    \includegraphics[width=.7\textwidth]{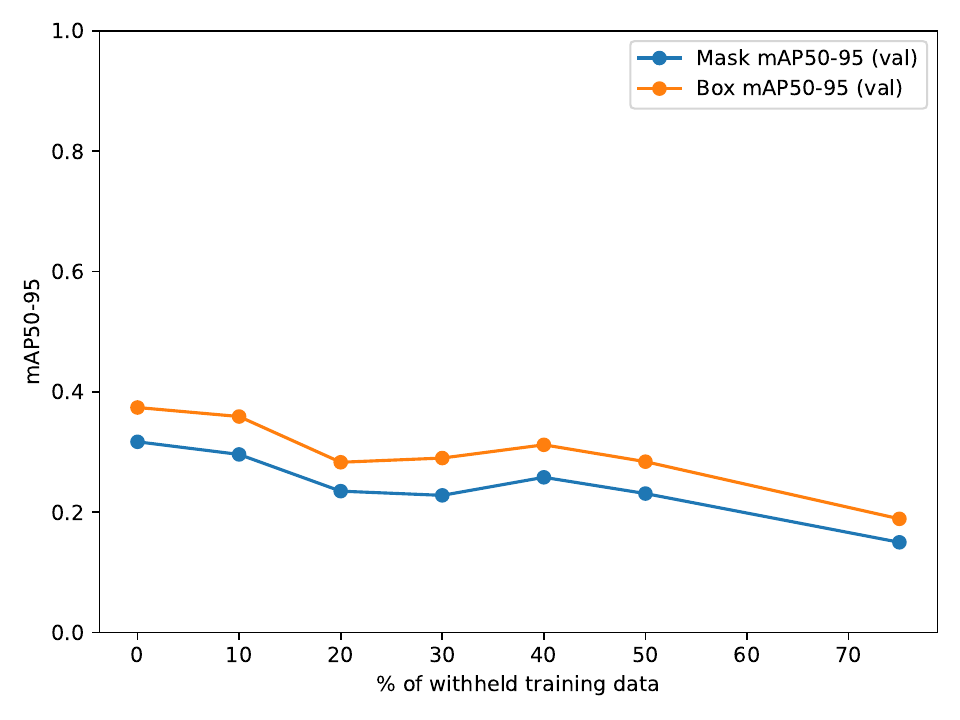}
    \caption{Best mask mAP$_{50-95}$ value (y-axis) achieved during training, measured on the val split, per portion of withheld training data (x-axis)}
    \label{fig:withhold_map}
\end{figure}
\begin{figure}[htbp]
    \centering
    \begin{subfigure}{.49\textwidth}
       \includegraphics[width=\textwidth]{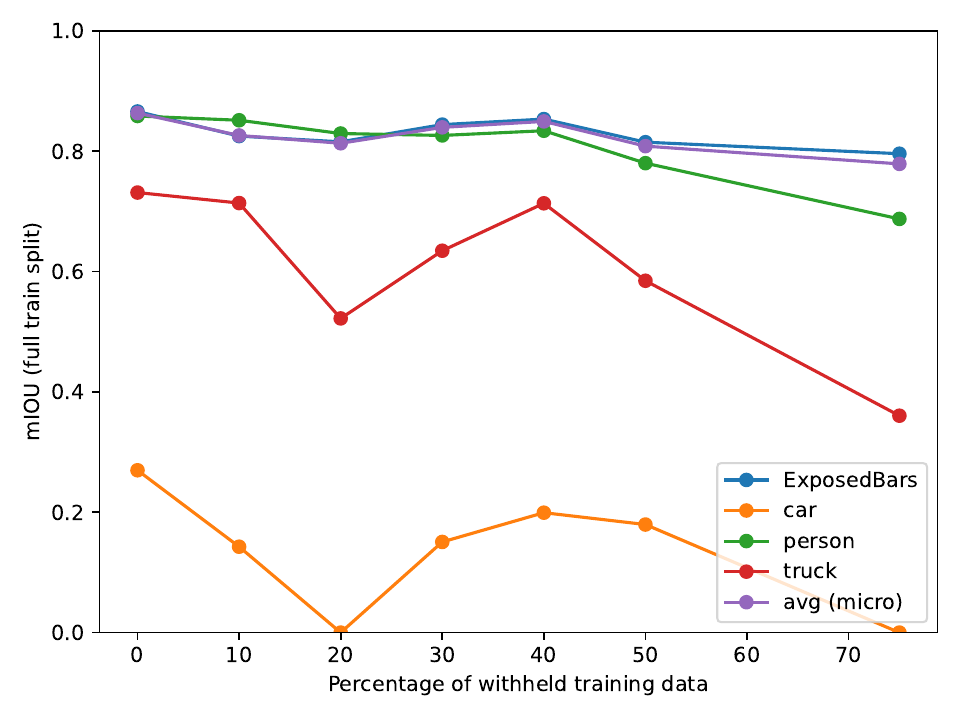} 
    \end{subfigure}
    \begin{subfigure}{.49\textwidth}
        \includegraphics[width=\textwidth]{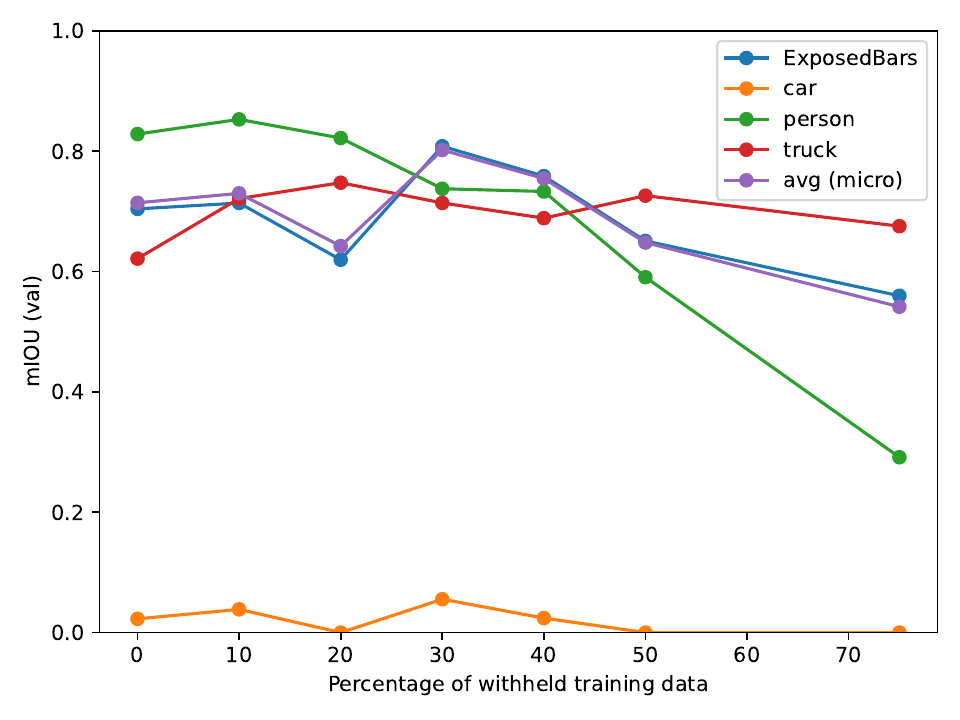}
    \end{subfigure}
    \caption{mIOU value (y-axis) of models trained on a percentage of withheld training data (x-axis), on the full train split (left) and val split (right). The purple line shows the micro average (class-agnostic mIOU), the other lines per-class mIOU.}
    \label{fig:withhold_miou}
\end{figure}

We train the YOLOv8L-seg instance segmentation model on the training data with \texttt{X} percent of it withheld. We start our training from pre-trained weights provided by the developers of YOLOv8L-seg and use default hyperparameters. Figures \ref{fig:withhold_map} and \ref{fig:withhold_miou} show the results of this experiment.


In Figure \ref{fig:withhold_map}, the blue line shows the best value of the mask mean Average Precision (mAP)$_{50-95}$ metric achieved during training on the y-axis, measured on the val split of the dataset at a certain percentage of withheld training data on the x-axis. The first point on the line shows the performance of the baseline with no withheld training data, corresponding to a value of 0.317. We observe a significant drop in performance at 20\% of withheld training data and again at 75\%. In Figure \ref{fig:withhold_miou}, we also report the mIOU metric per class and the percentage of withheld training data. The right plot (results on val split) shows a similar behavior, with the mIOU dropping significantly at 75\% of withheld training data. This is particularly visible for the \textit{person} class. We also observe a high discrepancy in performance for the \textit{car} class compared to the rest of the classes. However, a quick peek at the left plot of Figure \ref{fig:withhold_split_dist} reveals a negligible amount of instances for the \textit{car} class, explaining the poor performance of this class in general. Given the prominence of the \textit{ExposedBars} class in the dataset, the performance rather remains constant when withholding training data, which indicates that the data captures enough variation to train a robust model. The performance of other classes could for example be improved by using common segmentation datasets like MS COCO and including those in the training of the model.

\subsection{Effect of different labelling styles}
\label{subsec:train_without_anomaly}
Starting from the tagging process for anomalous samples as outlined in Section \ref{subsec:annotation_styles}, we analyze the influence of the presence of different annotation styles associated with visually similar samples. We withhold Langebro training samples tagged as anomalies, followed by a training of the YOLOv8L-seg model with the same training settings as the established baselines. To ensure a fair comparison, we also consider the amount of withheld data in the train and val split and use the results from Section \ref{subsec:data_efficiency} to have a second, more appropriate baseline to compare against. In general, when analyzing the effect of different labelling styles on model performance, we propose the following protocol for fair comparison:
\begin{enumerate}
    \item Perform experiments withholding certain amounts of training data while ensuring similar class distributions across the original and the smaller datasets.
    \item Identify deviating labelling styles in training and validation data and remove frames containing these. \begin{enumerate}
        \item Possibly, remove more samples to ensure a similar class distribution compared to the original training and validation data
    \end{enumerate}
    \item Re-train segmentation model on modified training data
    \item Evaluate model on modified validation data
    \item Compare obtained metrics to an appropriate baseline. Here, select the metrics of a model from Step 1 where the amount of training data is roughly the same as the amount of training data used in Step 3.
\end{enumerate}
In total, applying the rule outlined in Section \ref{subsec:annotation_styles} and following the previously mentioned protocol, we omit 150 samples from the train and 27 samples from the val split respectively. The effect of the instance counts on the train split are outlined in Table \ref{tab:instance_counts_anomalies}, for the val split in Table \ref{tab:instance_counts_anomalies_val}. Relatively speaking, the different annotation styles are more explained in the train split than in the val split, however, the differences are not too big, making it fair to conclude the general effect of this phenomenon in our dataset.
The results of this experiment are listed in Table \ref{tab:results_without_anomaly}. 

\begin{table}[htbp]
    \centering
        \caption{Instance counts of baseline train split and with anomalous samples removed}
   \begin{tabular}{lrrrrr}
   \hline
         & ExposedBars & person & truck & car  &\hspace{2em}Sum\\
         \hline
        baseline & 24,985& 9,105& 996& 180&35,266\\
        non-anomaly & 23,598& 9,047& 996& 180&33,821\vspace{1em}\\
 Difference to baseline& -1,387& -58& 0&0 &-1,445\\
 relative& -5.5 \%& -0.6 \%& 0& 0&-4.1 \%\\
 \hline
    \end{tabular} 
    \label{tab:instance_counts_anomalies}
\end{table}

\begin{table}[htbp]
    \centering
        \caption{Instance counts of baseline val split and with anomalous samples removed}
        \textcolor{black}{
        \begin{tabular}{lrrrrr}
        \hline
         & ExposedBars & person & truck & car  &\hspace{2em}Sum\\
        \hline
        baseline & 8,660& 3,225& 243 & 91  &12,219\\
        non-anomaly & 8,415& 3,133& 243 & 91  &11,882\vspace{1em}\\
 Difference to baseline& -245& -92& 0& 0&-337\\
 relative& -2.8 \%& -2.9 \%& 0& 0&-2.8 \%\\
 \hline
        \end{tabular}
        }
    \label{tab:instance_counts_anomalies_val}
\end{table}
The mIOU performance on both the train and val split has increased slightly. The same is valid for the val mask and box mAP$_{50-95}$.  The results of the training together with the small number of redacted samples and instances lead to the conclusion that the effect of these anomalous labels is minimal and can be disregarded. 

\begin{table}[htbp]
    \centering
\caption{Comparison between pre-trained baseline and pre-trained model with anomalous samples removed for training}
\label{tab:results_without_anomaly}
    \begin{tabular}{l>{\raggedleft\arraybackslash}p{0.27\textwidth}>{\raggedleft\arraybackslash}p{0.33\textwidth}}
    \hline
         &  YOLOv8L-seg (pre-trained)& YOLOv8L-seg (pre-trained, non-anomaly)\\
         \hline
         Checkpoint Epoch&  58& 18\\
         mIOU (train)&  0.681& 0.711\\
         mIOU (val)&  0.544 & 0.572 \\
 Mask mAP$_{50-95}$ (val)& 0.317 &0.304 \\
 Box mAP$_{50-95}$ (val)& 0.374 &0.385 \\
 \hline
    \end{tabular}
\end{table}
\par
\begin{figure*}[htbp]
    \centering
    \begin{subfigure}{.24\textwidth}
       \includegraphics[width=\textwidth]{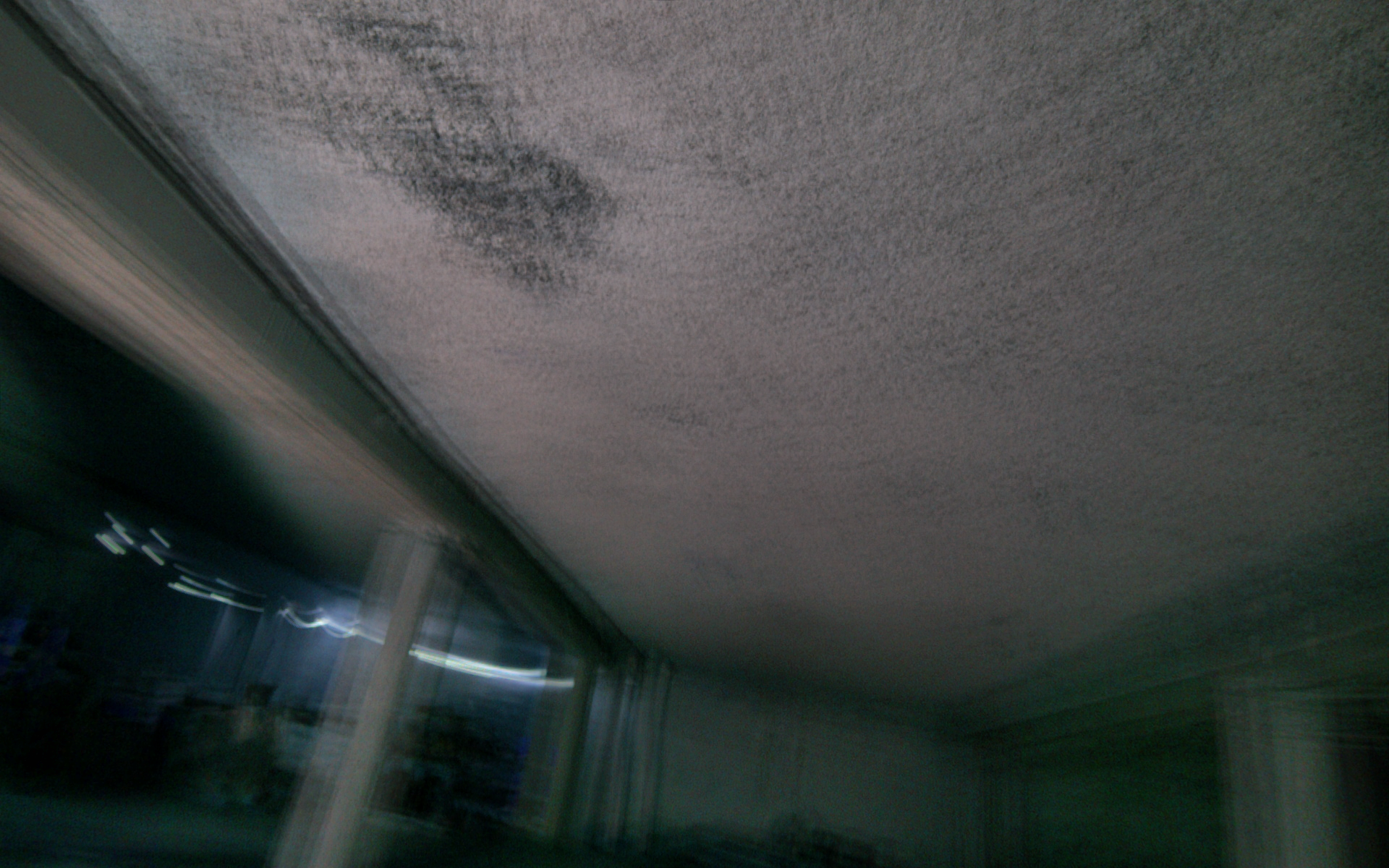} 
       \caption{Mean image}
    \end{subfigure}
    \vspace{1em}
    \begin{subfigure}{.24\textwidth}
       \includegraphics[width=\textwidth]{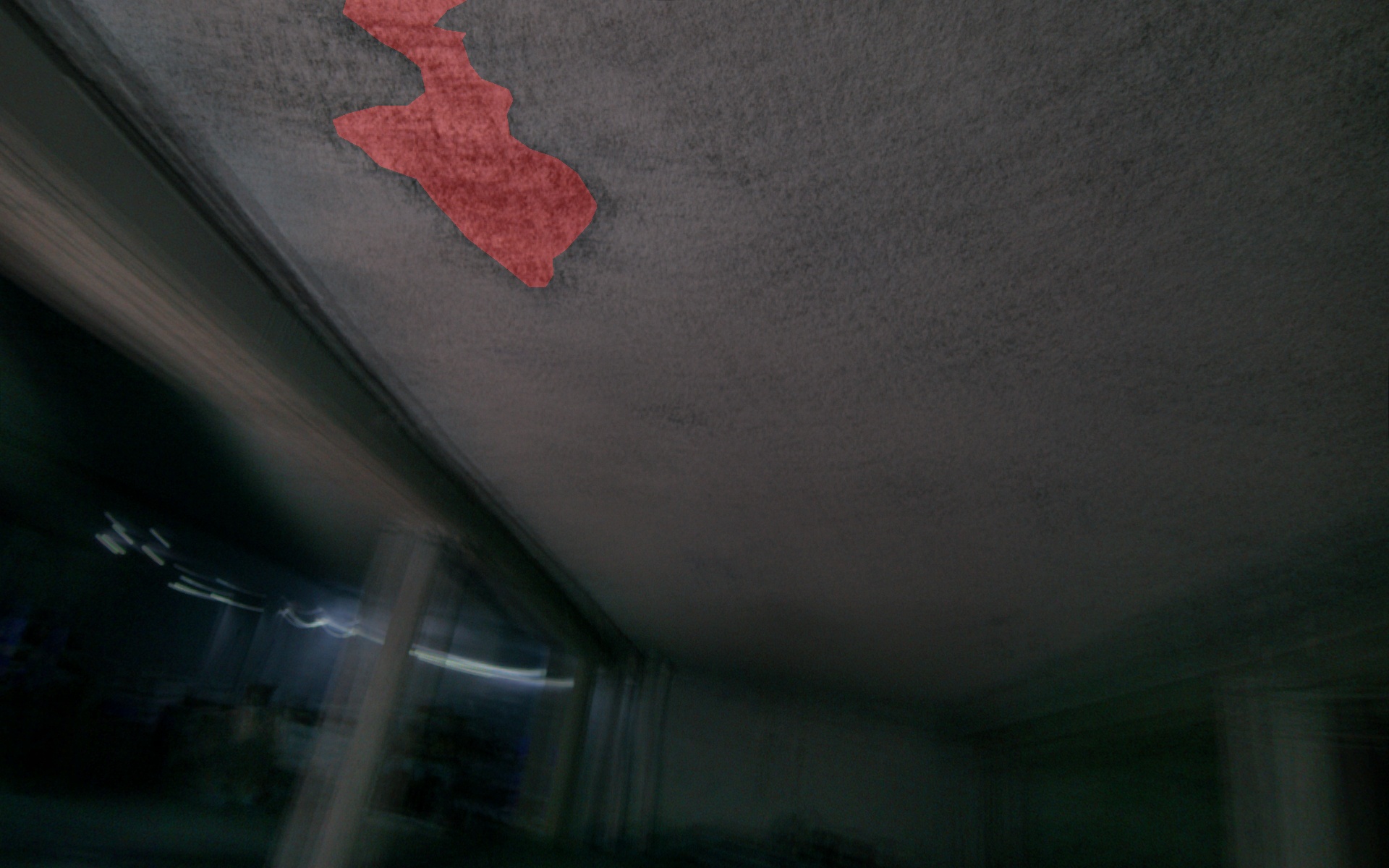} 
       \caption{Sample 0}
    \end{subfigure}
    \begin{subfigure}{.24\textwidth}
       \includegraphics[width=\textwidth]{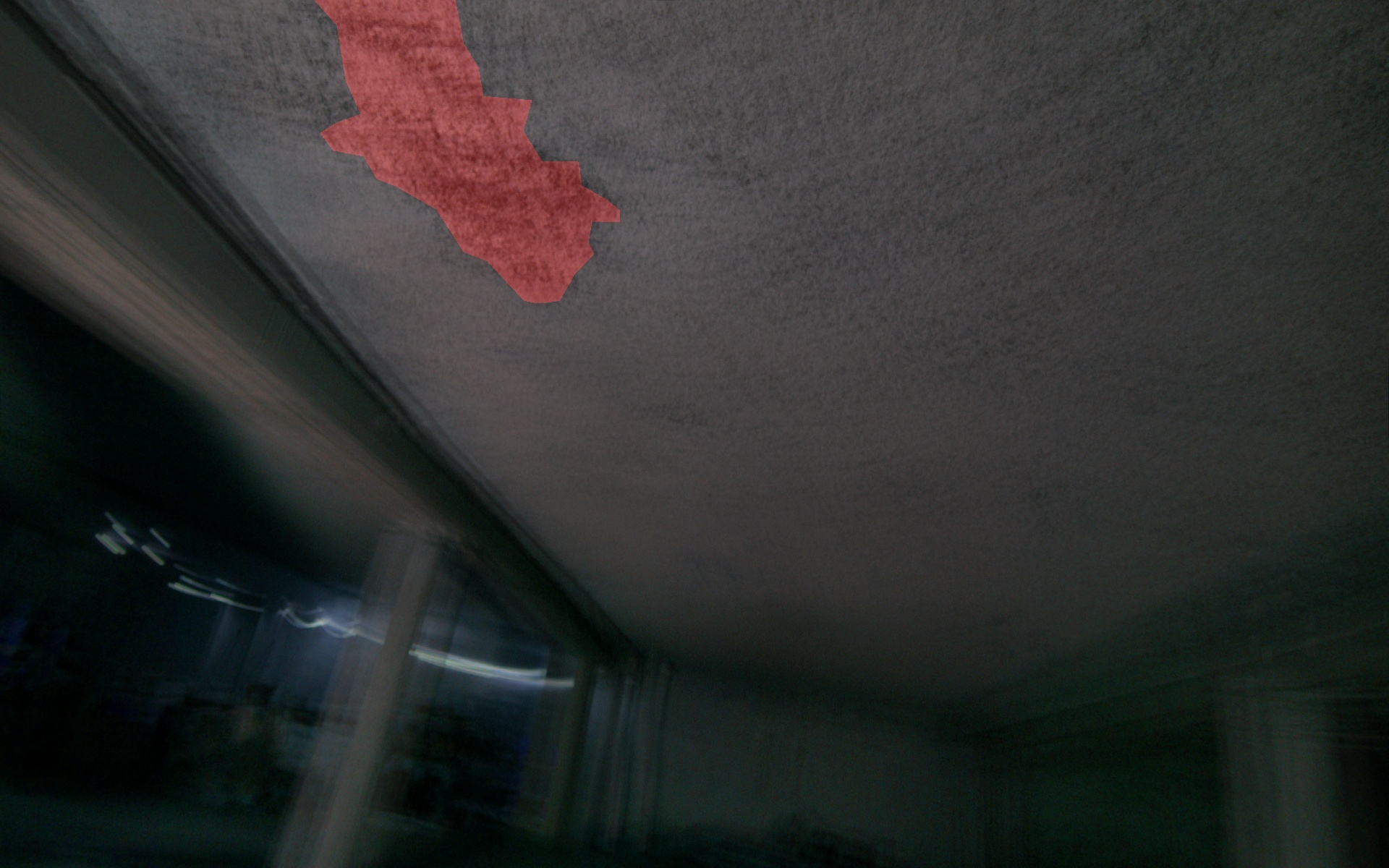} 
       \caption{Sample 1}
    \end{subfigure}
    \begin{subfigure}{.24\textwidth}
       \includegraphics[width=\textwidth]{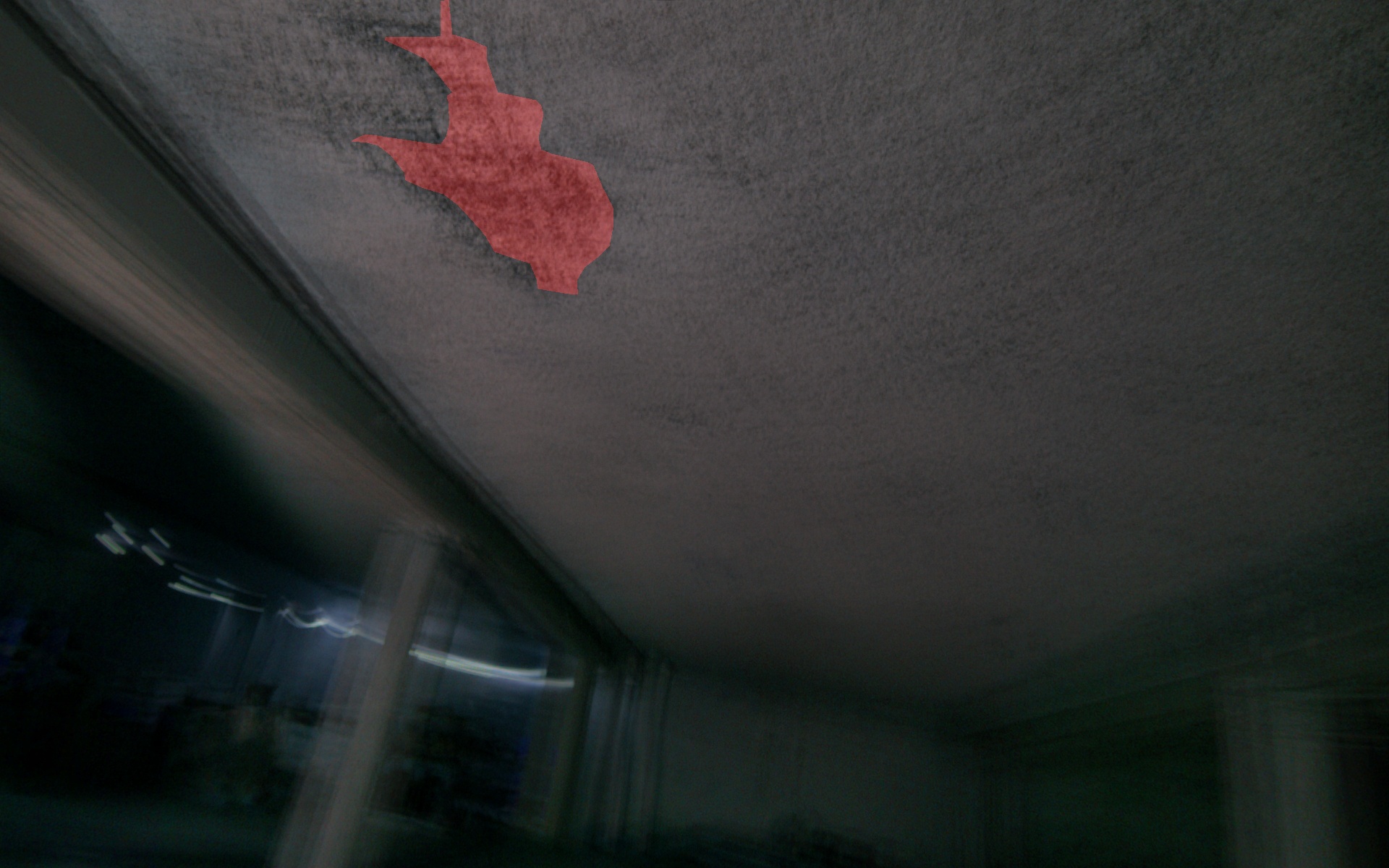} 
       \caption{Sample 2}
    \end{subfigure}
    \begin{subfigure}{.24\textwidth}
       \includegraphics[width=\textwidth]{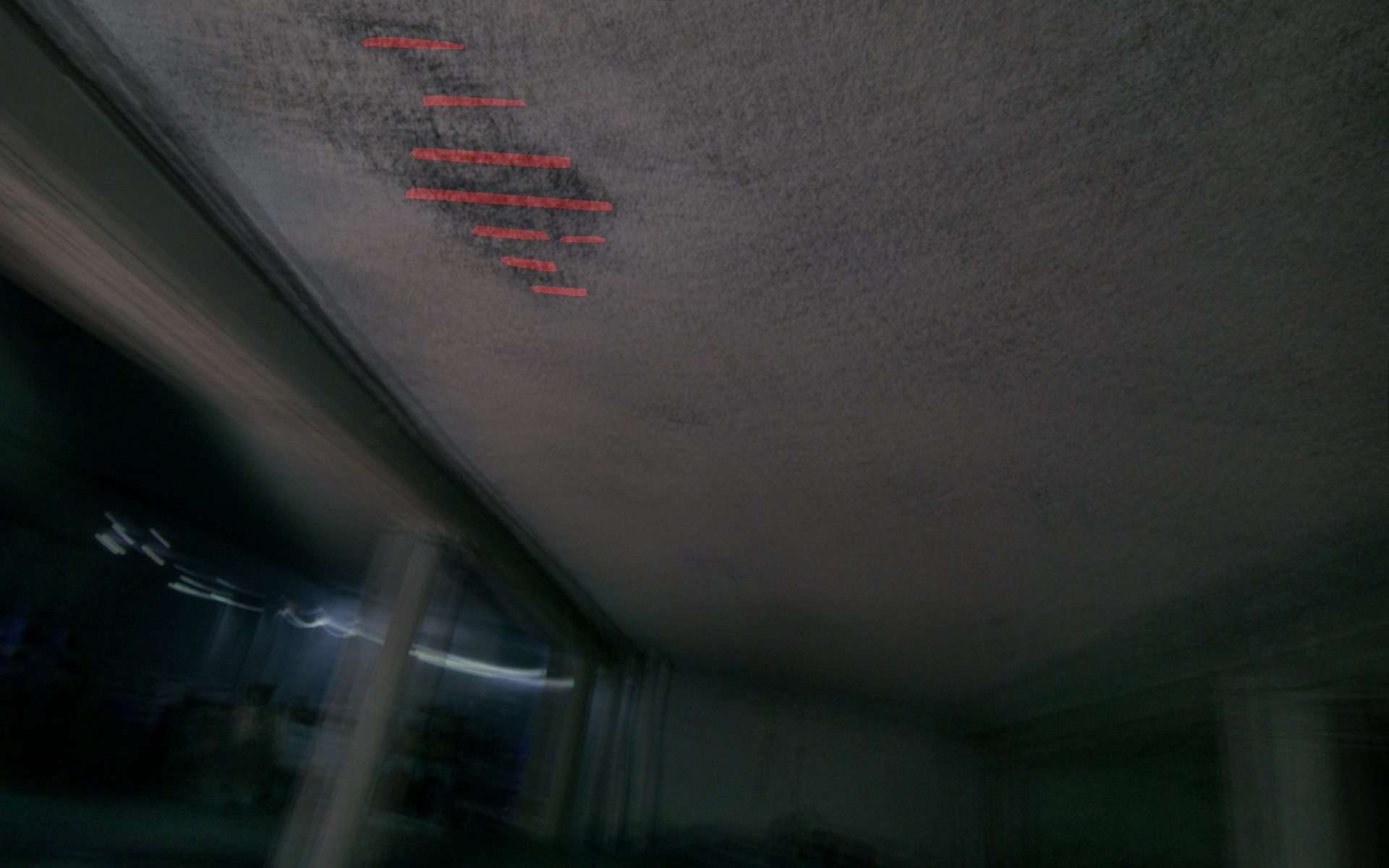} 
       \caption{Sample 3}
    \end{subfigure}
    \begin{subfigure}{.24\textwidth}
       \includegraphics[width=\textwidth]{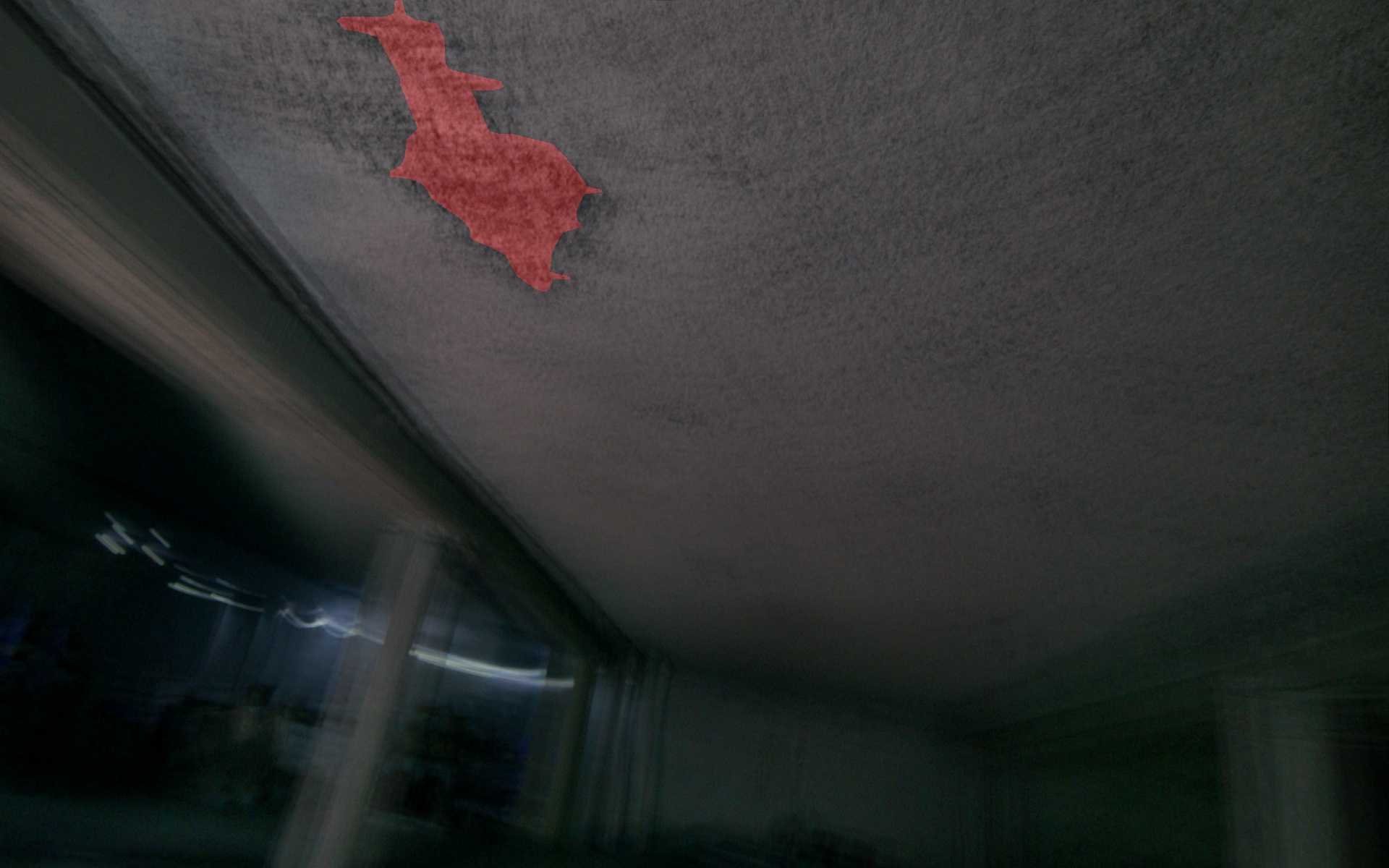} 
       \caption{Sample 4}
    \end{subfigure}
    \begin{subfigure}{.24\textwidth}
        \includegraphics[width=\textwidth]{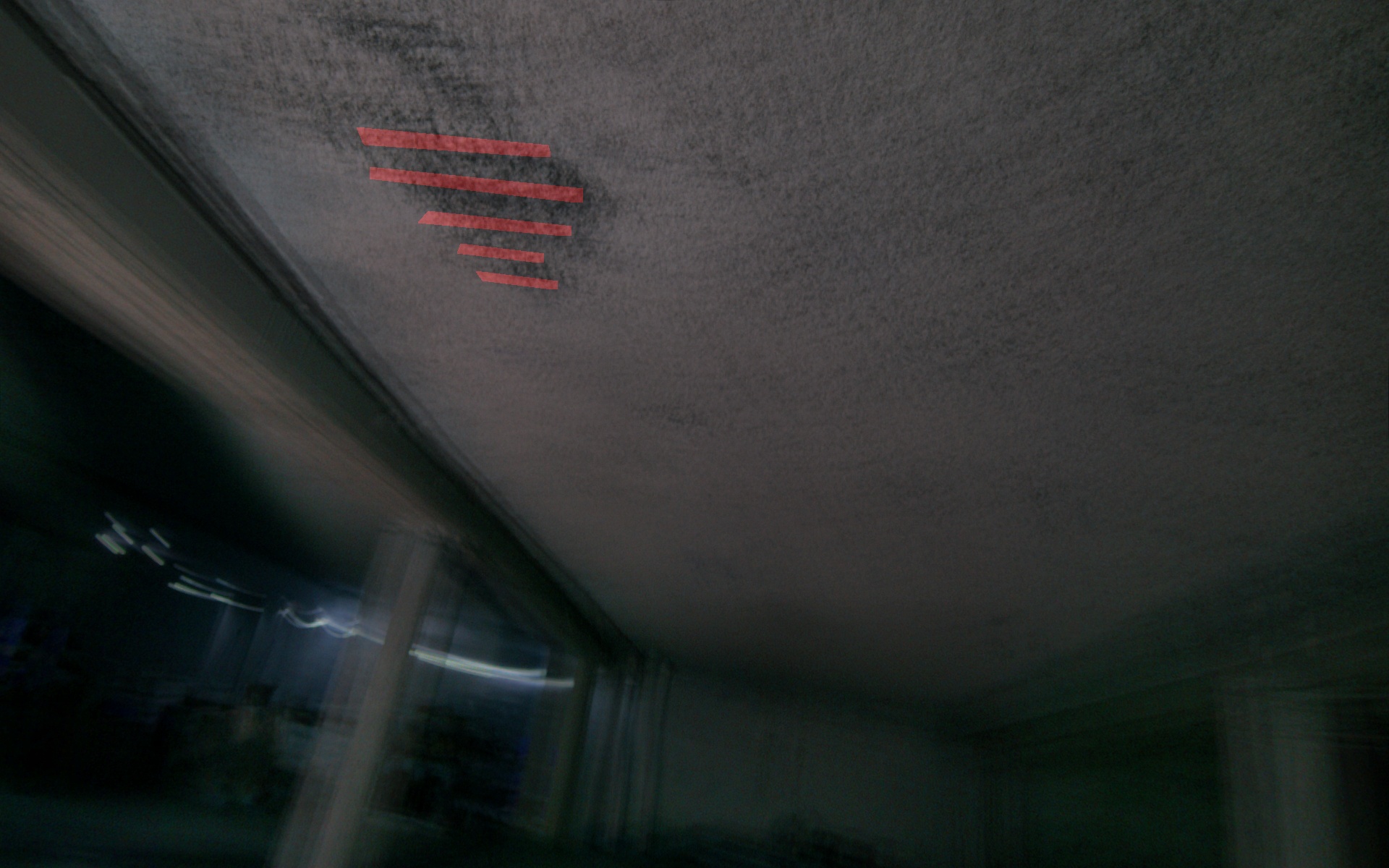}
        \caption{Sample 5}
    \end{subfigure}
    \begin{subfigure}{.24\textwidth}
       \includegraphics[width=\textwidth]{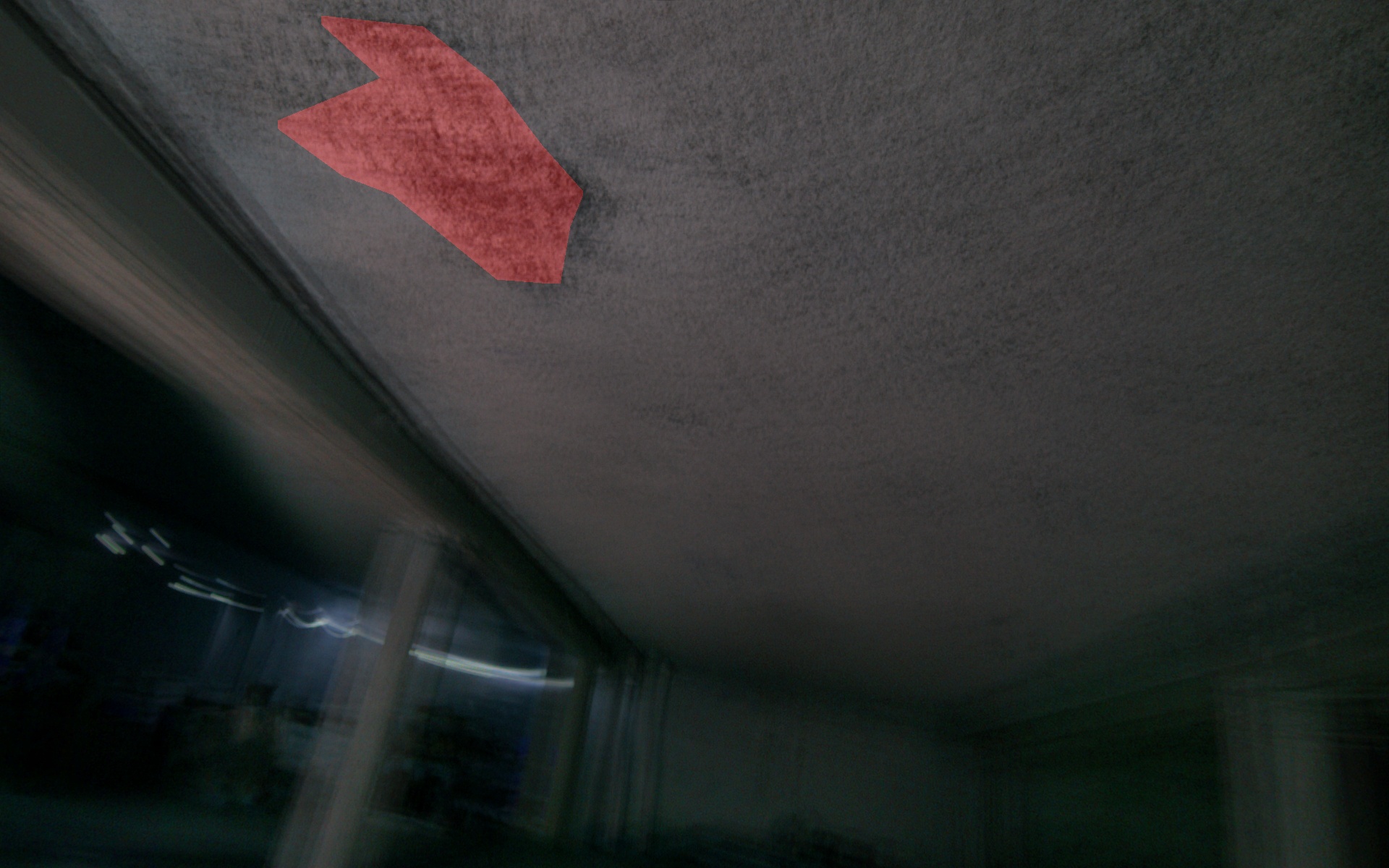} 
       \caption{Sample 6}
    \end{subfigure}
    \caption{Input data (labels in red) used to analyze gradient interactions between samples}
    \vspace{1em}
    \label{fig:mean_image_labels}
\end{figure*}

\begin{figure}[htbp]
    \centering
    \includegraphics[width=.7\textwidth]{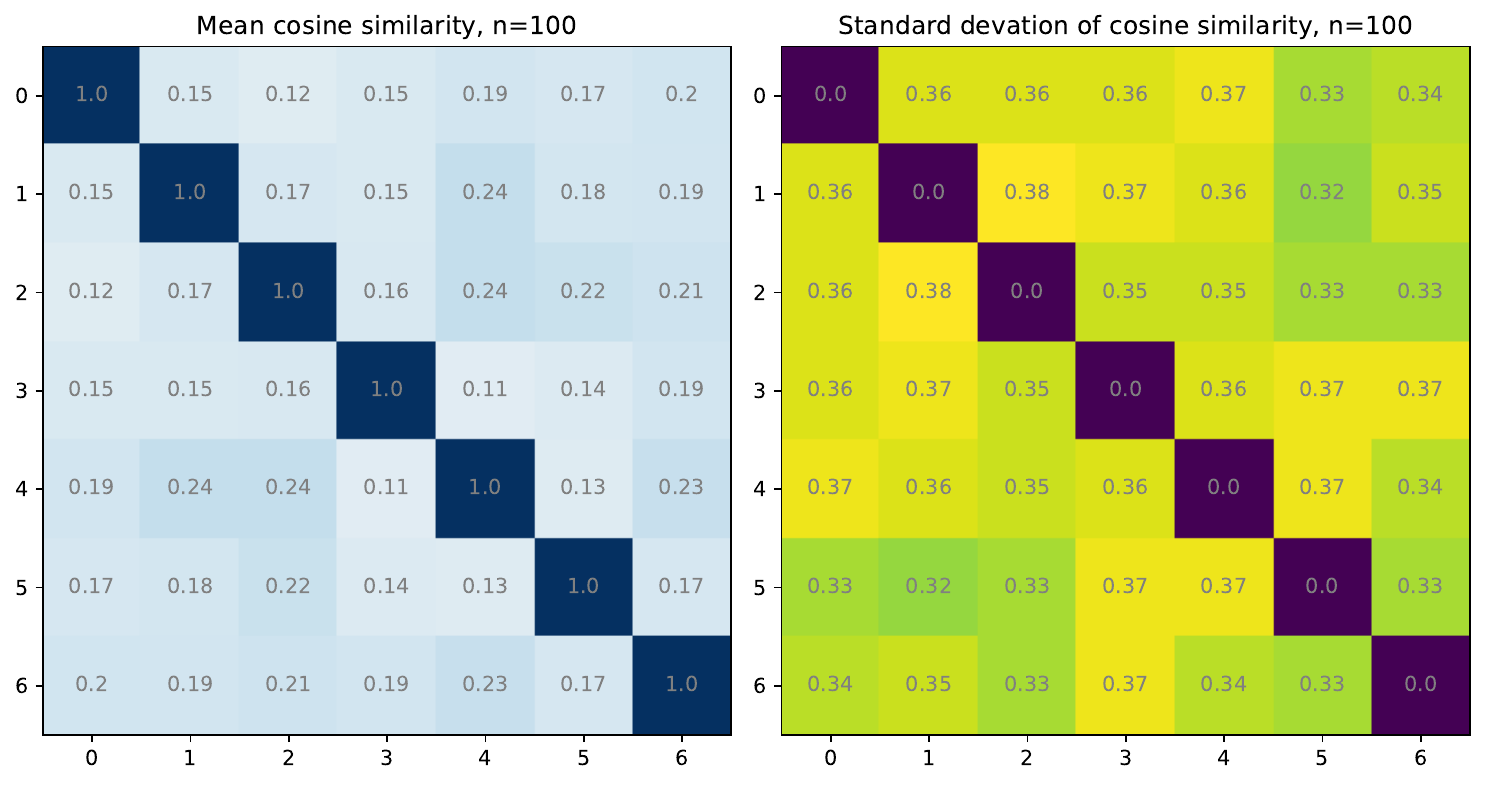}
    \caption{Cosine similarities between sample loss gradients, left mean cosine similarity, right standard deviation of cosine similarity. The tick labels are sample indices as listed in Figure \ref{fig:mean_image_labels}}
    \label{fig:gradient_interactions}
\end{figure}

 
We verify this by conducting another experiment, constructing a batch of seven consecutive samples (frame-label pairs). From the frames, we create a mean image to have a common input. We then pass this input through a DeepLabV3 model with a ResNet50 backbone, and use pre-trained weights (ImageNet) to create outputs. We then calculate the cross-entropy loss and evaluate the gradient of each loss with respect to all the model's parameters and investigate the cosine similarity of the respective gradients. Conflicting samples would then show a strong negative cosine similarity, whereas supporting samples would exhibit a strong positive cosine similarity. Figure \ref{fig:mean_image_labels} shows the data used for this experiment. Figure \ref{fig:gradient_interactions} shows the mean cosine similarity between the gradients for each sample loss. 

All values are positive, meaning that there is no gradient conflict among samples' loss gradients. There are also no values that strike out from the general picture. Therefore, this experiment supports the previously drawn conclusion that the influence of these different labelling styles is minimal and can be disregarded.

\section{Results}
In this section, we report results for the YOLOv8L-seg model trained from pre-trained weights on 100\% of the training data using default hyperparameters, as it was the best-performing model in our experiments conducted in Section \ref{sec:experiments}. For this final evaluation, we are also using the test split of the dataset. We will also identify common modes of failure and present ways to mitigate these. To verify the results, we have conducted additional verification runs with different seeds and report the results (min, mean, max) in Table \ref{tab:final_result}. Additionally, to reflect the robotic nature of the task, we have exported the best performing model of the verification runs to a TensorRT engine and tested the average FPS on an NVIDIA Jetson Xavier NX embedded GPU, with results reported in Table \ref{tab:jetson}, as high performance desktop GPUs are often not feasible on robotic platforms.
\begin{table}[htbp]
    \centering
    \caption{Final results, descriptive statistics based on four verification runs with equal configuration and different seeds}
    \label{tab:final_result}
    \begin{tabular}{>{\raggedright\arraybackslash}p{0.15\linewidth}>{\raggedleft\arraybackslash}p{0.05\linewidth}>{\raggedleft\arraybackslash}p{0.05\linewidth}>{\raggedleft\arraybackslash}p{0.07\linewidth}>{\raggedleft\arraybackslash}p{0.05\linewidth}>{\raggedleft\arraybackslash}p{0.05\linewidth}>{\raggedleft\arraybackslash}p{0.07\linewidth}>{\raggedleft\arraybackslash}p{0.05\linewidth}>{\raggedleft\arraybackslash}p{0.05\linewidth}>{\raggedleft\arraybackslash}p{0.05\linewidth}} 
    \hline
         &  \multicolumn{3}{c}{\textbf{min}}&  \multicolumn{3}{c}{\textbf{mean}}& \multicolumn{3}{c}{\textbf{max}}\\ 
 & train& val& test& train& val& test& train& val&test \\
 \hline
         mIOU &   0.660 & 0.555 &0.329 &  0.684 & 0.571 &0.345 & 0.702 & 0.590 &0.367 \\ 
         Mask mAP$_{50-95}$ &  0.487  & 0.287  &0.169  &  0.514  & 0.292  &0.174  & 0.532  & 0.298  &0.184  \\ 
 Mask mean precision &  0.718 & 0.582 &0.348 &  0.762 & 0.626 &0.373 & 0.803 & 0.692 &0.390 \\ 
 Mask mean recall&  0.616 & 0.305 &0.201 &  0.634 & 0.315 &0.209 & 0.653 & 0.322 &0.219 \\ 
 \hline
    \end{tabular}
\end{table}

\begin{table}[htbp]
    \centering
    \caption{Results of the best performing model of Table \ref{tab:final_result}, when converted to TensorRT engines of different weight types (integer, half precision and full precision). FPS are reported on Jetson Xavier NX in 20W hexacore mode}
    \label{tab:jetson}
    \begin{tabular}{>{\raggedright\arraybackslash}p{0.15\linewidth}>{\raggedleft\arraybackslash}p{0.05\linewidth}>{\raggedleft\arraybackslash}p{0.05\linewidth}>{\raggedleft\arraybackslash}p{0.07\linewidth}>{\raggedleft\arraybackslash}p{0.05\linewidth}>{\raggedleft\arraybackslash}p{0.05\linewidth}>{\raggedleft\arraybackslash}p{0.07\linewidth}>{\raggedleft\arraybackslash}p{0.05\linewidth}>{\raggedleft\arraybackslash}p{0.05\linewidth}>{\raggedleft\arraybackslash}p{0.05\linewidth}} 
    \hline
         &  \multicolumn{3}{c}{\texttt{INT8}}&  \multicolumn{3}{c}{\texttt{FP16}}& \multicolumn{3}{c}{\texttt{FP32}}\\ 
 & train& val& test& train& val& test& train& val&test \\
 mIOU& 0.507& 0.259& 0.146& 0.723& 0.571& 0.322& 0.724& 0.571&0.322\\
 \scriptsize$\Delta\%$&\scriptsize -27.8&\scriptsize -56.1&\scriptsize -60.2&\scriptsize 3.0&\scriptsize -3.2&\scriptsize -12.3&\scriptsize 3.1&\scriptsize -3.2&\scriptsize -12.3\\
         Mask mAP$_{50-95}$ &  0.367& 0.264& 0.121&  0.518& 0.296& 0.171& 0.518& 0.297& 0.171\\
 \scriptsize$\Delta\%$&\scriptsize -31.0&\scriptsize -11.4&\scriptsize -34.2&\scriptsize -2.6&\scriptsize -0.7&\scriptsize -7.1&\scriptsize -2.6&\scriptsize -0.3&\scriptsize-7.1\\ 
 Mask mean precision &  0.729& 0.701&0.434&   0.756& 0.653& 0.374&  0.754& 0.653&0.375\\
 \scriptsize$\Delta\%$&\scriptsize -9.2&\scriptsize 1.3&\scriptsize 11.3&\scriptsize -5.9&\scriptsize -5.6&\scriptsize -4.1&\scriptsize -6.1&\scriptsize -5.6&\scriptsize-3.8\\ 
 Mask mean recall&  0.337& 0.068&0.070&  0.637& 0.327&0.208& 0.64& 0.327&0.207\\
 \scriptsize$\Delta\%$&\scriptsize -48.4&\scriptsize -78.9&\scriptsize -68.0&\scriptsize -2.5&\scriptsize 1.6&\scriptsize -5.0&\scriptsize -2.0&\scriptsize 1.6&\scriptsize-5.5\\ 
 \hline
 Avg. FPS& \multicolumn{3}{c}{20.92}& \multicolumn{3}{c}{12.79}& \multicolumn{3}{c}{4.61}\\
 \hline
    \end{tabular}
\end{table}

For all metrics, the performance is roughly stable within a split, with only small fluctuations.  When measuring the same metrics on the different splits of the dataset, the performance naturally decreases. This is a sign that the model is overfitting and is usually alleviated by introducing more variation in the training data, accompanied by regularization techniques. Table \ref{tab:final_result} also shows that the drop from val to test is not as drastic as from train to val, indicating that the validation split is representative of the test split as well.
\par
When analyzing mean mask precision and recall, we see that the model performance is heavily impacted by a significant drop in recall. While the drop in precision is about half a magnitude when going from the train split to the test split, it is roughly two-thirds for the recall.  This means that the model often fails to localize the objects in general. The root cause of this is usually an insufficient number of instances in the dataset. Since Table \ref{tab:final_result} shows macro metrics, it is imperative to recall that our dataset is imbalanced and does not have a lot of instances for classes other than the \textit{ExposedBars} category. Thus, it makes sense to separately calculate these metrics solely for this category, which is listed in Table \ref{tab:final_result_eb_only}. We can see that the \textit{ExposedBars} category performs better than the macro average in terms of mIOU -- the test mIOU shows an improvement of almost a magnitude. The mAP$_{50-95}$ on the test split is slightly higher than the macro average, probably due to the increased precision for the \textit{ExposedBars} category. Compared to the macro average mean value, the mean mask precision is almost double as high when solely considering the \textit{ExposedBars} category. The mean mask recall is roughly on par with the macro average mean value. 
When converting the model to a TensorRT engine for deployment on an embedded GPU such as the NVIDIA Jetson Xavier NX in our case, performance drops are to be expected and also observed, as indicated in Table \ref{tab:jetson}. The difference in segmentation metrics between the full floating-point precision (\texttt{FP32}) and half floating-point precision (\texttt{FP16}) are minimal, however, there are significant performance drops when using integer quantization (\texttt{INT8}) of the model weights. The performance drops of \texttt{FP16} and \texttt{FP32} are minimal, with a relative decrease of 3 \% in validation mIOU, while achieving a frame rate of 12.8 and 4.6 FPS respectively. We therefore recommend using the \texttt{FP16} engine as a good compromise between model accuracy and frame rate.
\begin{table}[tb]
    \centering
    \caption{Final results, descriptive statistics based on four verification runs with equal configuration and different seeds, on the \textit{ExposedBars} category only}
    \label{tab:final_result_eb_only}
    \begin{tabular}{>{\raggedright\arraybackslash}p{0.15\linewidth}>{\raggedleft\arraybackslash}p{0.05\linewidth}>{\raggedleft\arraybackslash}p{0.05\linewidth}>{\raggedleft\arraybackslash}p{0.07\linewidth}>{\raggedleft\arraybackslash}p{0.05\linewidth}>{\raggedleft\arraybackslash}p{0.05\linewidth}>{\raggedleft\arraybackslash}p{0.07\linewidth}>{\raggedleft\arraybackslash}p{0.05\linewidth}>{\raggedleft\arraybackslash}p{0.05\linewidth}>{\raggedleft\arraybackslash}p{0.07\linewidth}} 
    \hline
         &  \multicolumn{3}{c}{\textbf{min}}&  \multicolumn{3}{c}{\textbf{mean}}& \multicolumn{3}{c}{\textbf{max}}\\ 
 & train& val& test& train& val& test& train& val&test \\
 \hline
         mIOU &  0.884 & 0.607 &0.564 &  0.889 & 0.675 &0.617 & 0.893 & 0.716 &0.718 \\ 
         Mask mAP$_{50-95}$ &  0.590 & 0.182 &0.257 &  0.618 & 0.194 &0.267 & 0.638 & 0.211 &0.273 \\ 
 Mask mean precision &  0.787 & 0.470 &0.669 &  0.822 & 0.512 &0.679 & 0.854 & 0.533 &0.691 \\ 
 Mask mean recall&  0.775 & 0.161 &0.199 &  0.783 & 0.176 &0.210 & 0.793 & 0.196 &0.218 \\ 
 \hline
    \end{tabular}
\end{table}
Since we used the YOLOv8L-seg model, an instance segmentation model, we also have detected bounding boxes at our disposal. We therefore use these to analyze the failure modes and feed them into the TIDE tool \cite{bolya_tide_2020}, a tool for identifying object detection errors. The TIDE tool categorizes detection errors into five main categories and assigns them a $\Delta$mAP value, which reports the increase in mAP that one could expect if all errors of a type are resolved. The categories are as follows:
\begin{itemize}
    \item \textbf{Classification errors}, i.e., properly localized bounding box, but wrong class label
    \item \textbf{Localization errors}: a bounding box has an IOU with a detection bigger than a background IOU threshold, but lower than a foreground IOU threshold
    \item \textbf{Duplicate detection errors}: Multiple predicted bounding box have an IOU bigger than the foreground IOU threshold with the ground truth bounding box
    \item \textbf{Background errors}: A false positive detection, i.e., a predicted bounding box has an IOU smaller than the background threshold with a ground truth box
    \item \textbf{Missed detection}: There's no matched predicted bounding box for a ground truth bounding box (false negative)
\begin{figure}[htbp]
    \centering
    \begin{subfigure}{.3\textwidth}
        \includegraphics[width=\textwidth]{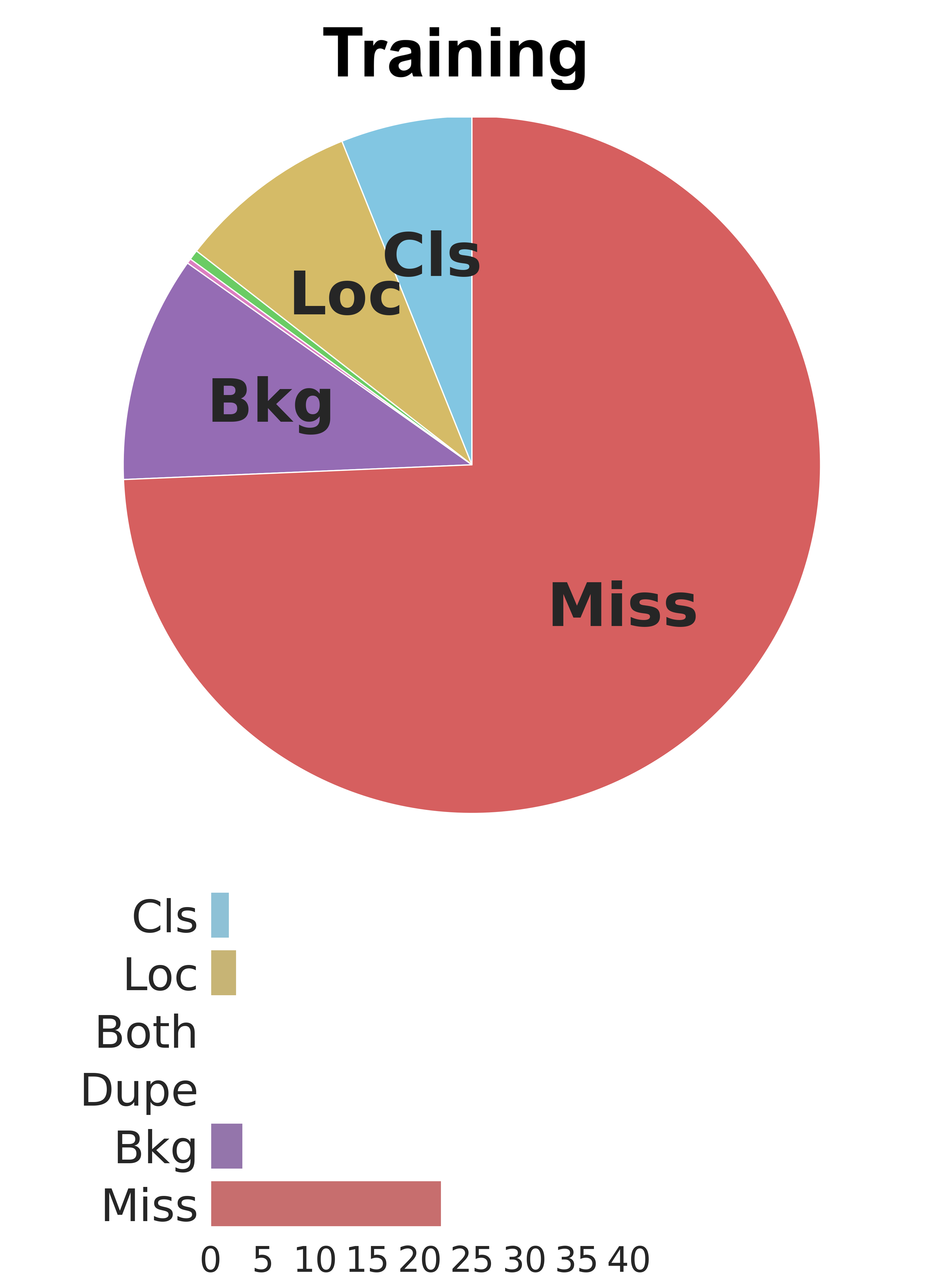}
    \end{subfigure}
    \begin{subfigure}{.29\textwidth}
        \includegraphics[width=\textwidth]{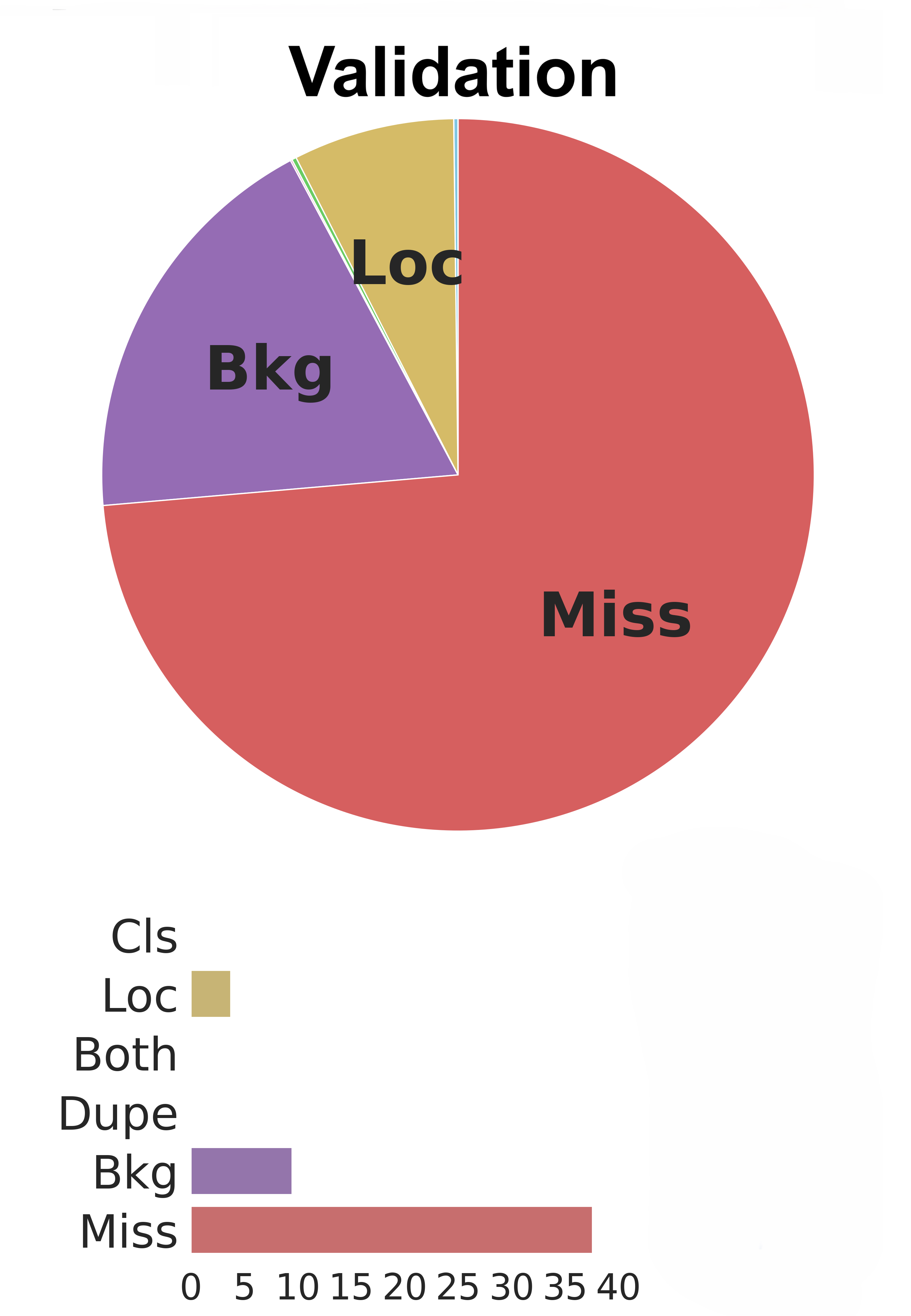}
    \end{subfigure}
       \begin{subfigure}{.29\textwidth}
        \includegraphics[width=\textwidth]{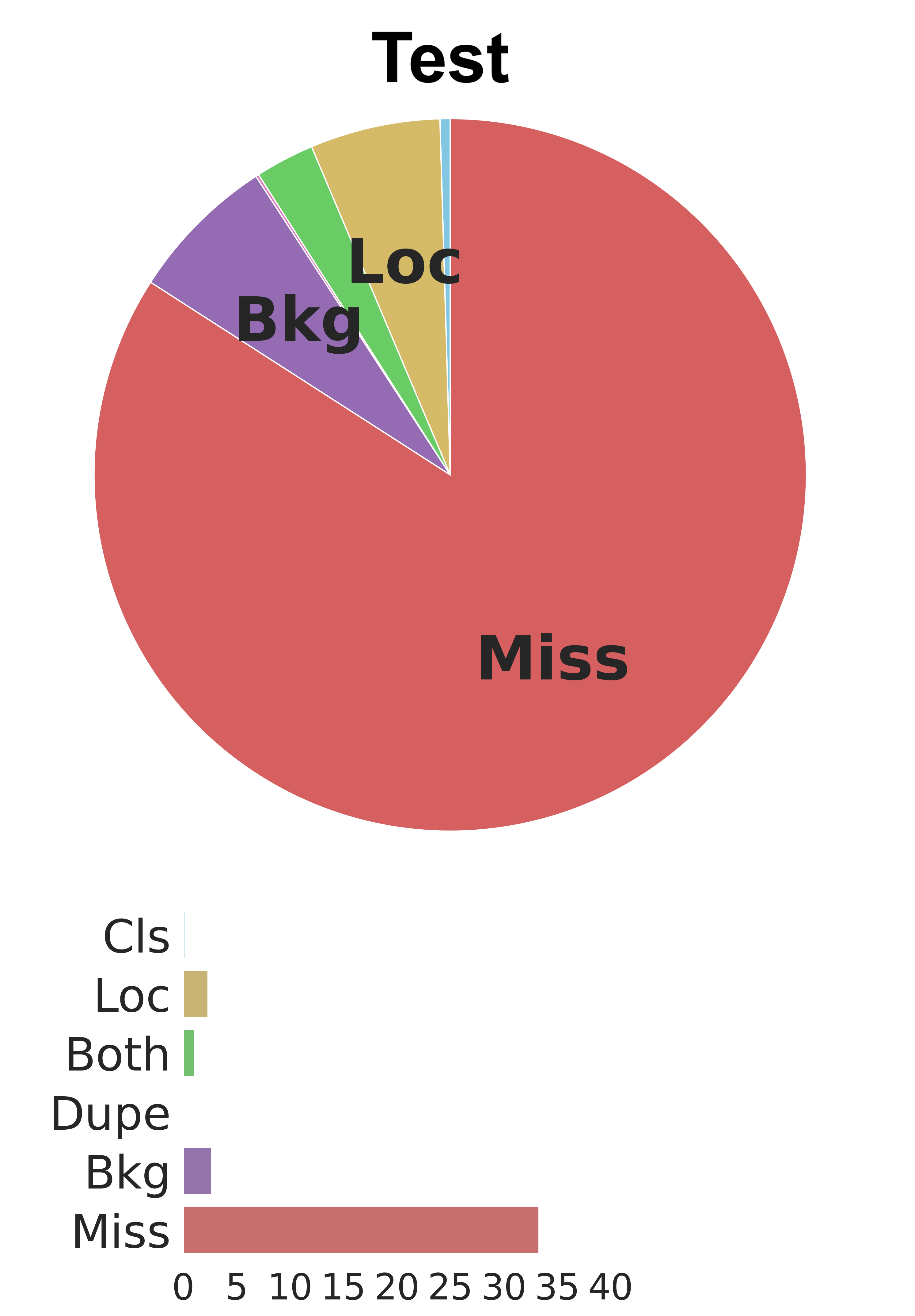}
    \end{subfigure}
    \caption{Error analysis of bounding box predictions, per split. Output from TIDE tool}
    \label{fig:tide_analysis}
\end{figure}
\end{itemize}
Figure \ref{fig:tide_analysis} shows the results of the bounding box error analysis conducted with the TIDE tool. Throughout the splits, one big source of errors are missed detections, i.e., the model fails to produce a prediction for some ground truth instance. After this type, false positive detections are also a significant source of errors, i.e., the model recognizes unimportant backgrounds as an object. Localization and classification errors are, compared to the other error types, minimal. We can also see that $\Delta$mAP value for missed detections is almost double compared to the train split, which is in line with the mask mean recall values reported in Tables \ref{tab:final_result} and \ref{tab:final_result_eb_only}.
\par
To conclude, the results show that our trained model is precise and show good segmentation performance for the \textit{ExposedBars} class, but struggles to generalize across splits. This indicates that our dataset has qualitative annotations, but still has room for improvement in terms of the amount of data and variety captured.
\section{Discussion}
Our study was aimed at introducing a new dataset for reinforced concrete construction, with a focus on public availability, contrary to most other similar works. Furthermore, it was desired to create a dataset with spatiotemporal relationships, reflecting the realistic robotic application of this dataset. 
The following summarizes our key findings, which will be elaborated later in this section:
\begin{itemize}
    \item We have found that the performance of the segmentation model YOLOv8L-seg decreases with more and more withheld training data. 
    \item We have showed a positive influence of pre-training on model performance, even when trained on unrelated, general data like the MS COCO dataset.
    \item We have showcased labelling inconsistencies in our dataset, and found that removing these inconsistencies doesn't alter model performance significantly.
    \item We have also showed that our data has a certain center bias, i.e., the mask centroids are concentrated in the middle of the image. 
    \item Our data shows high discrepancies in model performance when evaluated on a per-class level, with the \textit{car} class showing low performance in terms of mIOU.
    \item An error analysis of our model revealed that false negative errors are by far the biggest error mode. 
    \item On an embedded GPU, using half-floating point precision engines yields a good compromise between model accuracy and frame rate.
\end{itemize}
\par
Our experiment showing the decrease in model performance with more and more withheld training data suggests also the opposite, i.e. model performance increases with more and more data availability. Given the lack of publicly available datasets for construction applications, we identify this property as one of the main factors why training well-performing deep learning-based models for construction applications is challenging. We therefore urge future work to make data publicly accessible to alleviate this barrier.
\par
Because our data has a center bias in its annotations, the superiority of the YOLOv8L-seg over DeepLabV3 and U-Net can easily be explained, since YOLOv8L-seg uses a sophisticated set of image augmentations, making the model more robust to these peculiarities. Furthermore, our dataset is biased towards the \textit{ExposedBars} class, with a very minimal amount of instances of the \textit{car} class. This explains the discrepancies in per-class performance. This limitation can be overcome by including data from other datasets that include these classes.
\par
\begin{figure}[htbp]
    \centering
    \includegraphics[width=.98\textwidth]{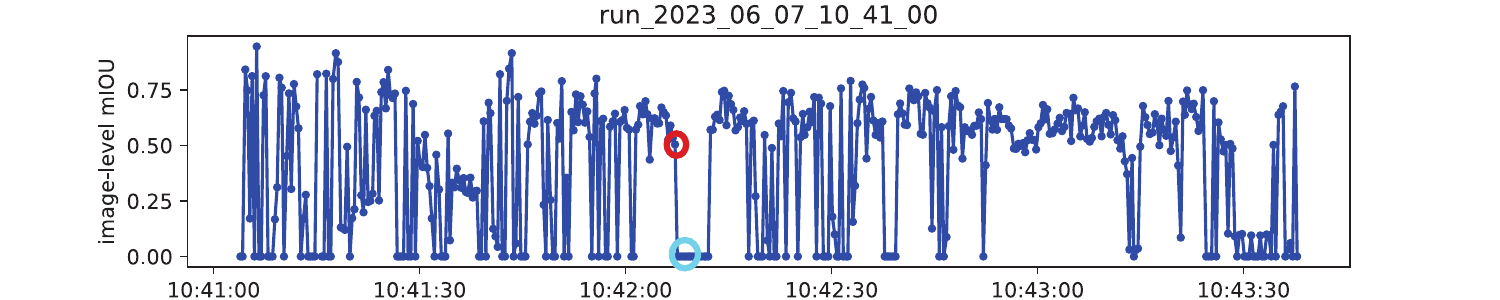}\\
    \vspace{1em}
    \begin{subfigure}[t]{.48\textwidth}
       \includegraphics[width=\textwidth]{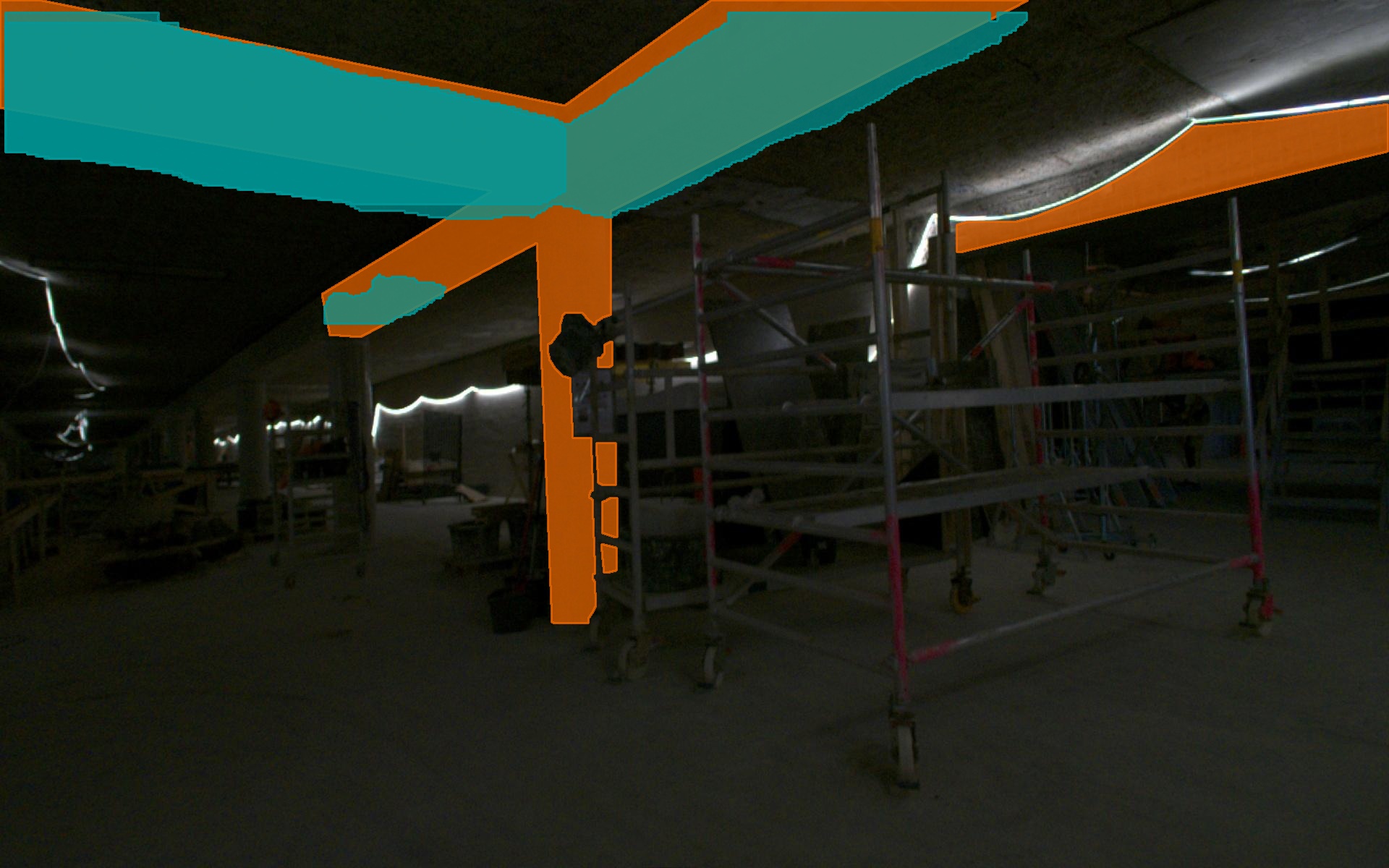}
       \caption{Reference frame with existing prediction, referring to the red circle in the upper plot}
    \end{subfigure}
    \begin{subfigure}[t]{.48\textwidth}
        \includegraphics[width=\textwidth]{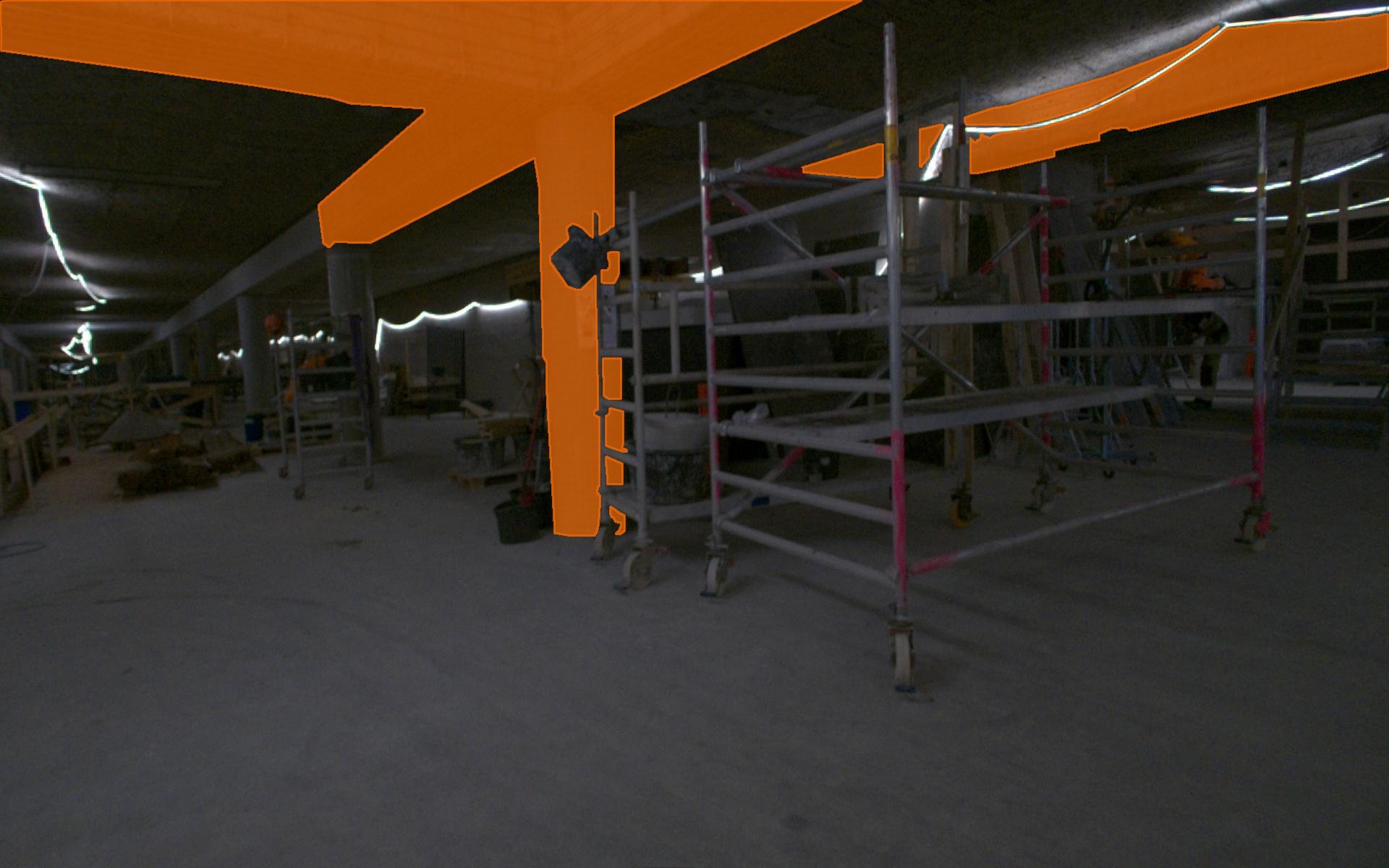}
        \caption{Frame with missing prediction of same object, referring to the cyan circle in the upper plot}
    \end{subfigure}
    \caption{Illustration of model instabilities. Model instabilities often include a significant drop in mIOU, followed by a sudden increase in mIOU. Predictions are in cyan, ground truth annotations in orange}
    \label{fig:mIOU_time}
\end{figure}

The false negative errors are a clear limitation of our model. We have further analyzed this phenomenon by investigating the development of mIOU values over time in a scene and could see that some false negatives are just model instabilities, i.e. the model fails to produce a prediction for an object previously detected. Figure \ref{fig:mIOU_time} illustrates this. This problem has already been discussed and analyzed in \cite{schmidt_towards_2024}. We stipulate that these errors are due to minor perturbations of the input resulting in big distances in feature space, thus failing to produce predictions. We therefore recommend the usage of mask stabilization techniques, as outlined in \cite{schmidt_towards_2024}. This also supports the approach of having a spatiotemporal dataset, since models are applied in real life, where one can leverage these dependencies to increase system performance.\par
While our research showed minimal influences of different labelling styles on label performance, we acknowledge that this insight cannot be generalized given the low number of samples in our dataset showing this phenomenon. However, given the positive gradient interaction presented in Section \ref{subsec:train_without_anomaly}, we recommend analyzing this phenomenon in other settings to verify our findings, if applicable. Furthermore, we stipulate that deep learning-based segmentation models in construction are a good use case for probabilistic segmentation models, such as Probabilistic U-Net \cite{kohl_2018}, as those try to model different, valid expert opinions on how to segment a certain object. Further research should investigate the application of such models in a construction context.
\section{Conclusion}
This paper presented a dataset consisting of temporally ordered RGB images with segmentation labels for reinforced concrete construction. The dataset is, as opposed to most datasets in construction, made publicly available to boost research in the area and contains 14,805 images in total, bigger than most task-specific datasets that focus on the construction sector. We have carried out an extensive analysis of our dataset and specifically analyzed variations in annotation styles. Furthermore, we have established and compared three baseline segmentation models: U-Net, DeepLabV3, and YOLOv8L-seg. Our study revealed that the YOLOv8L-seg model trained from pre-trained weights is superior in performance compared to U-Net and DeepLabV3. Moreover, we have explored the effects of withholding training data while maintaining an equal class distribution and have found that the performance of the model increases with more available training data. In addition, we have explored the influence of different annotation styles on the model performance by removing anomalous annotations from the training data and have found no significant change. In addition, we implemented the model on an embedded GPU and found a good compromise between model accuracy and frame rate using \texttt{FP16} precision. Lastly, our final evaluation showed that the model is precise in predicting areas with exposed reinforcement bars, but fails to generalize and recognize these areas in scenes that deviate from those depicted in the train split.
\par
Our study concludes that there is still a significant lack of CV data for construction tasks. Unlike other sectors, where digitization has been adopted swiftly in the past decades, construction has been traditionally slow to adopt digital technologies. This has resulted in low data availability and, thus, a challenging environment to develop deep learning-based solutions to be used in autonomous robotic systems in construction. These robotic systems will be a necessity in the future, as the construction sector is facing labor shortages in the upcoming years. We think the community will benefit from more publicly available datasets and a common data lake, where researchers will upload data along with annotations, even if annotations styles within a category are vastly different, as our study showed the minimal impact of this phenomenon. 
\par
We hope that our work will motivate other researchers to contribute their data and thus work towards a large data pool, paving the way for better model performance for computer vision tasks in construction.

\vspace{6pt}

\section*{Acknowledgements} This work has been funded and supported by the EU Horizon Europe project ``RobétArmé” under the Grant Agreement 101058731. We extend our gratitude to Christiansen \& Essenbaek A/S (CEAS) and Markus Schmidtke, for sharing their domain expertise and organizing access to their construction sites for data-capturing purposes.
\section*{Data Statement}
The self-collected scenes of the dataset can be found on DTU Data \cite{schmidt_andersen_casaslorenzo_gasconbononad_2024}. The code to initiate the dataset including the dataset management tool can be found on GitHub in our repository\footnote{\url{https://github.com/DTU-PAS/ConRebSeg}}.
\section*{Author contributions: CRediT}
\textbf{Lazaros Nalpantidis}: Funding acquisition, Supervision, Writing -- Review and Editing

\bibliography{references.bib, additional_references.bib}
\end{document}